\documentclass{article}

\PassOptionsToPackage{numbers, compress}{natbib}

\usepackage[preprint]{neurips_2025}

\makeatletter
\def\@noticestring{}
\makeatother

\usepackage{wrapfig}
\usepackage[utf8]{inputenc} 
\usepackage[T1]{fontenc}    

\usepackage{lmodern}

\usepackage{hyperref}       
\usepackage{url}            
\usepackage{booktabs}       
\usepackage{amsfonts}       
\usepackage{nicefrac}       
\usepackage{microtype}      
\usepackage{xcolor}         
\usepackage{float}          
\usepackage{placeins}       

\definecolor{linkgray}{HTML}{716C6F}

\hypersetup{
    colorlinks=true,
    linkcolor=linkgray,
    citecolor=linkgray,
    urlcolor=linkgray,
    pdfborder={0 0 0}
}

\usepackage{amsmath,amsfonts,bm}

\def\eqref#1{equation~\ref{#1}}

\def\1{\bm{1}}

\DeclareMathAlphabet{\mathsfit}{\encodingdefault}{\sfdefault}{m}{sl}
\SetMathAlphabet{\mathsfit}{bold}{\encodingdefault}{\sfdefault}{bx}{n}

\usepackage{subcaption}
\usepackage{xspace}
\usepackage{multirow}

\usepackage{graphicx}
\usepackage{esvect}
\usepackage{enumitem}
\usepackage{mathrsfs}

\usepackage[normalem]{ulem}
\useunder{\uline}{\ul}{}

\usepackage{amssymb}
\usepackage{mathtools}
\usepackage{amsthm}
\usepackage{cleveref}

\usepackage[most]{tcolorbox}

\newtcolorbox{promptbox}[1][]{
  colback=gray!10,
  colframe=gray!50,
  coltitle=black,
  boxrule=0.8pt,
  arc=6pt,
  left=6pt,
  right=6pt,
  top=6pt,
  bottom=6pt,
  fonttitle=\bfseries,
  title=#1,
  fontupper=\ttfamily\raggedright
}

\newcommand{\HATifiedModelSmall}{Llama-3.1-8B-TFree-HAT\xspace}
\newcommand{\HATifiedModelLarge}{Llama-3.1-70B-TFree-HAT\xspace}
\newcommand{\HATpretrained}{Llama-TFree-HAT-Pretrained\xspace}
\newcommand{\coo}{\ensuremath{\mathrm{CO_2}}}
\newtheorem{defn}{Definition}

\title{A Family of LLMs Liberated from \\Static Vocabularies}

\begin{document}

\vspace{0.5cm}

\begin{center}
    \includegraphics[width=0.2\textwidth]{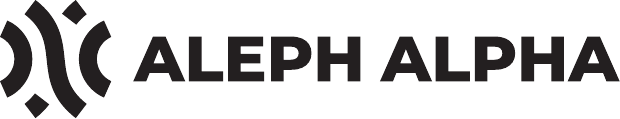}
\end{center}

\author{}
\maketitle

\vspace{-1.75cm}

\begin{center}
    Aleph Alpha Research \\
    \vspace{0.35cm}
    \tiny{A detailed author list can be found in the appendix of this paper.}
\end{center}

\vspace{0.25cm}

\begin{figure}[!b]
    \centering
    \vspace{-1cm}
    \includegraphics[width=1\textwidth]{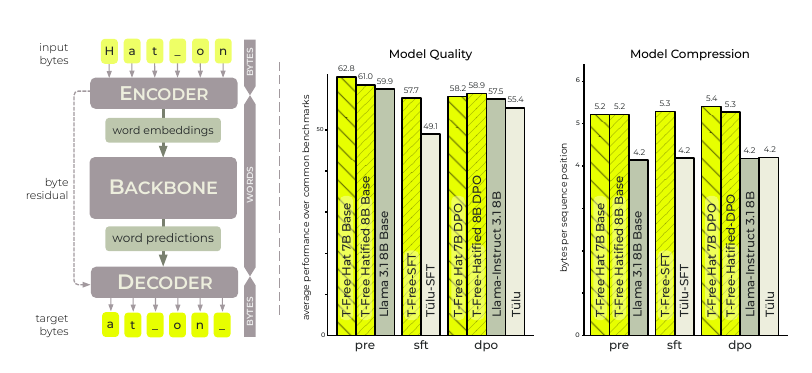}
    \caption{(Left) The HAT architecture has three components: an encoder, backbone, and decoder, each implemented as a transformer. A full overview can be found in Figure~\ref{fig:full-architecture}, while the encoder and decoder are detailed in Figures~\ref{fig:encoder-cross-attn} and \ref{fig:decoder-cross-attn}, respectively. (Right) Average performance and compression for \HATifiedModelSmall on benchmarks detailed in §\ref{sec:performance}.}
    \label{fig:arch_and_results_average}
\end{figure}

\begin{abstract}

Tokenization is a central component of natural language processing in current large language models (LLMs), enabling models to convert raw text into processable units. Although learned tokenizers are widely adopted, they exhibit notable limitations, including their large, fixed vocabulary sizes and poor adaptability to new domains or languages. We present a family of models with up to 70 billion parameters based on the hierarchical autoregressive transformer (HAT) architecture. In HAT, an encoder transformer aggregates bytes into word embeddings and then feeds them to the backbone, a classical autoregressive transformer. The outputs of the backbone are then cross-attended by the decoder and converted back into bytes. We show that we can reuse available pre-trained models by converting the Llama 3.1 8B and 70B models into the HAT architecture: \HATifiedModelSmall and \HATifiedModelLarge are byte-level models whose encoder and decoder are trained from scratch, but where we adapt the pre-trained Llama backbone, i.e., the transformer blocks with the embedding matrix and head removed, to handle word embeddings instead of the original tokens. We also provide a 7B HAT model, trained entirely from scratch on nearly 4 trillion words, \HATpretrained. The HAT architecture improves text compression by reducing the number of required sequence positions and enhances robustness to intra-word variations, e.g., spelling differences. Through pre-training, as well as subsequent supervised fine-tuning and direct preference optimization in English and German, we show strong proficiency in both languages, improving on the original Llama 3.1 in most benchmarks. We release our models (including 200 pre-training checkpoints) on \href{https://huggingface.co/Aleph-Alpha}{Hugging Face}.

\end{abstract}

\section{Introduction} \label{sec:intro}

We introduce a series of language models, one pre-trained entirely from scratch and others based on Llama 3.1 8B or 70B \cite{grattafiori2024llama3herdmodels}, augmented with a novel tokenizer replacement: the hierarchical autoregressive transformer (HAT) architecture, originally described by Neitemeier et al.\ \cite{neitemeierhierarchical} and further extended in this work. HAT integrates byte-level encoding and decoding with a word-level transformer backbone\footnote{By `backbone' we mean the transformer blocks with the embedding matrix and head removed, to handle the word embeddings instead of the subword tokens.}. This hierarchical structure provides potential advantages: (1) increased robustness to prompt perturbations; and (2) improved adaptability to new data, e.g., domains and languages, through continued training.

A foundational innovation in our work is the extension of a `tokenizer-free' (T-Free) approach to LLM training and inference by splitting raw byte data into variable-length chunks rather than mapping inputs to a fixed vocabulary. While this method could technically be viewed as a form of tokenization, we argue in §\ref{subsection:tokenization} that it differs meaningfully in practice, especially in how it avoids large embedding tables and allows models to exploit similarity between chunks of bytes. To clarify this distinction, we provide formal definitions of both classical tokenization and our alternative perspective.

We pre- and post-trained our models in English and German on curated corpora. To encourage helpfulness and instruction adherence, we performed direct-preference optimization (DPO). This makes the model more suitable for real-world applications, reducing the likelihood of unnecessary refusals. Notably, these models demonstrate strong performance in German, while also outperforming the original Llama 3.1 models on many English-language benchmarks.

It is important to note that we did not optimize these models for code generation or mathematical reasoning; accordingly, we do not extensively evaluate them on those tasks. While the HAT architecture provides intrinsic efficiency benefits, real-world inference speed is significantly influenced by the quality of the inference implementation. We report our work on incorporating vLLM inference for our models in §\ref{sec:inference}.

\paragraph{Contributions}

In this report:
\begin{enumerate}
    \item we show that pre-trained tokenizer-free approaches can be competitive with tokenized equivalents (including when trained on one-third of the total pre-training data budget of Llama-3.1 models);
    \item we demonstrate that a pre-trained model's tokenizer can be successfully replaced with a tokenizer-free approach -- a method which we here dub \textit{HATification} -- while improving downstream performance and compression ratios and, in the case of the 8B model, reducing the total number of model parameters by more than 10\%;
    \item we outline a pipeline for LLM-development from data curation and pre-training to post-training and inference;
    \item we make our models publicly available to the research community in order to contribute to further advancements in tokenizer-free large language models (LLMs). Specifically, we release checkpoints (i) for our base models\footnote{\url{https://huggingface.co/Aleph-Alpha/tfree-hat-pretrained-7b-base} \\ and \url{https://huggingface.co/Aleph-Alpha/llama-3_1-8b-tfree-hat-base}} (including 200 pre-training checkpoints over training for future study and use by the community), (ii) with supervised fine-tuning\footnote{\url{https://huggingface.co/Aleph-Alpha/llama-3_1-8b-tfree-hat-sft} \\ and \url{https://huggingface.co/Aleph-Alpha/llama-3_1-70b-tfree-hat-sft}}, and (iii) with direct-preference optimization\footnote{\url{https://huggingface.co/Aleph-Alpha/llama-tfree-hat-pretrained-7b-dpo} \\ and \url{https://huggingface.co/Aleph-Alpha/llama-3_1-8b-tfree-hat-dpo}}; and
    \item we introduce our evaluation framework\footnote{\url{https://github.com/Aleph-Alpha-Research/eval-framework}} and vLLM inference contributions for tokenizer-free model inference\footnote{\url{https://github.com/Aleph-Alpha/vllm}}, both of which we make available under Apache 2.0.
\end{enumerate}

\section{General Overview} \label{sec:overview}

Our models adopt and extend the HAT architecture \cite{neitemeierhierarchical}, which enhances byte-level language modeling with intermediate word-level representations. The architecture consists of three components---encoder, backbone, and decoder---each implemented as an autoregressive transformer, together with connector layers between components. The encoder operates on UTF-8 byte sequences using local causal attention and aggregates byte-level embeddings into word-level embeddings via cross-attention with learned queries. These embeddings are processed by the backbone, a standard causal transformer operating at word resolution, to produce contextual word representations. The decoder then generates next-byte predictions using both byte-level context and cross-attention to the backbone's word-level outputs. This architecture enables efficient long-context modeling by compressing input through word-level abstraction while preserving fine-grained detail at the byte level---which we hypothesize also improves multilingual adaptation.

At the start of the training process, we randomly initialized the encoder, decoder, and connector layers. For \HATpretrained, we pre-trained from scratch (i.e. randomly initialized the backbone), whereas for \HATifiedModelSmall and \HATifiedModelLarge we initialized the backbone from pre-trained Llama 3.1 weights \cite{grattafiori2024llama3herdmodels}. In the first pre-training phase, we trained our models using a next-byte prediction objective with sequences up to 3,500 words\footnote{We also set an upper bound of 28,000 bytes, i.e., an average of 8 characters per word.}. In this first phase, \HATpretrained was trained for nearly 4T words, \HATifiedModelSmall for 134B words, and \HATifiedModelLarge for 108B words; for comparison, we note that Llama 3.1 models were trained with 15T tokens, which is approximately 12T words.

We then continued training with longer sequences of words, emphasizing longer documents to adapt to extended context\footnote{There is no fixed byte sequence length as words have different byte lengths, but we enforce an upper bound on byte sequence length of 28,000 bytes during initial pre-training and 262,144 bytes for the long context adaptation.}. In this second phase, \HATpretrained was trained on sequences up to 32,768 words for 10.5B words, \HATifiedModelSmall on sequences up to 32,768 words for 20B words, and \HATifiedModelLarge on sequences up to 16,000 words for 10.2B words. For \HATifiedModelSmall and \HATifiedModelLarge, during initial pre-training, we kept the backbone frozen for the first 2,000 steps and then trained with a reduced learning rate, while the encoder and decoder followed a warmup-stable-decay schedule. For long-context adaptation, we kept most parameters frozen except query and key projections in attention layers, using a smaller learning rate and longer data sequences, while maintaining a similar data mix \citep{fu2024dataengineeringscalinglanguage}.

We evolved the inference implementation of HAT from a simple HuggingFace-based prototype to a production-ready system built on vLLM \citep{kwon2023efficient}, adapted for batched serving. Integrating HAT into vLLM surfaced architectural challenges stemming from HAT's hierarchical design, particularly its dual-sequence processing and variable-length byte-level generation. These posed significant obstacles for batching, requiring modifications to the scheduling strategy to balance backbone utilization and latency while managing asynchronous generation patterns within a batch. A key innovation was the management of dual key-value (KV) caches -- one for byte-level and one for word-level sequences -- necessitating careful memory coordination due to their asymmetric and interdependent resource demands. Throughout our implementation, we prioritized compatibility with vLLM's core to minimize invasive changes, enabling scalable deployment without compromising the hierarchical model's unique execution semantics.

Post-training involved supervised fine-tuning (SFT) on a diverse mix of $\sim 2$M samples, including synthetic responses (especially in German) generated and filtered using strong open models, as well as human-written data. Finally, we performed alignment using DPO, with careful filtering of preference pairs to improve helpfulness and safety. Averaged over our evaluations, our models achieve strong scores relative to Llama 3.1 equivalents.

\section{Model}
We detail the architecture of our models in Section~\ref{subsection:architecture}, the approach for splitting byte sequences into words in Section~\ref{subsection:word_splitter}, the difference between fixed tokenization and our approach in Section~\ref{subsection:tokenization}, and infrastructure and code optimizations in Section~\ref{subsection:code_optimizations}.

\begin{figure}[tb]
    \centering
    \includegraphics[scale=0.25]{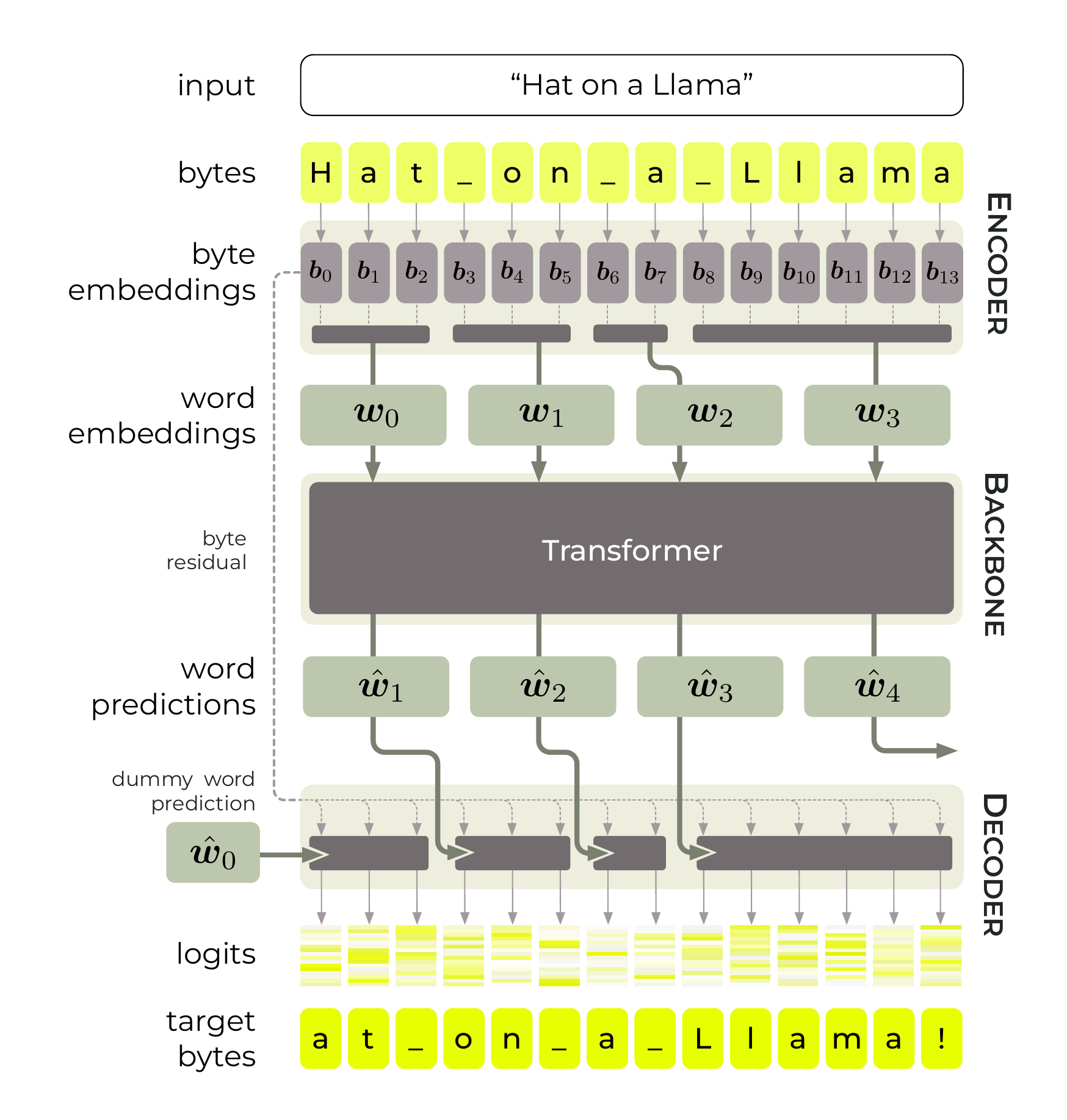}
    \caption{Overview of our model architecture. The encoder and decoder are detailed in Figures~\ref{fig:encoder-cross-attn} and \ref{fig:decoder-cross-attn} respectively. The encoder processes the input text, producing word embeddings $\bm{w}_k$, which are then processed by the backbone to produce next word predictions $\hat{\bm{w}}_{k+1}$. The decoder uses these predictions along with encoder's byte-level outputs $\bm{b}$ to generate byte-level logits.}
    \label{fig:full-architecture}
\end{figure}

\begin{figure}[tb]
    \centering
    \begin{subfigure}[t]{0.35\textwidth}
        \centering
        \includegraphics[scale=0.25]{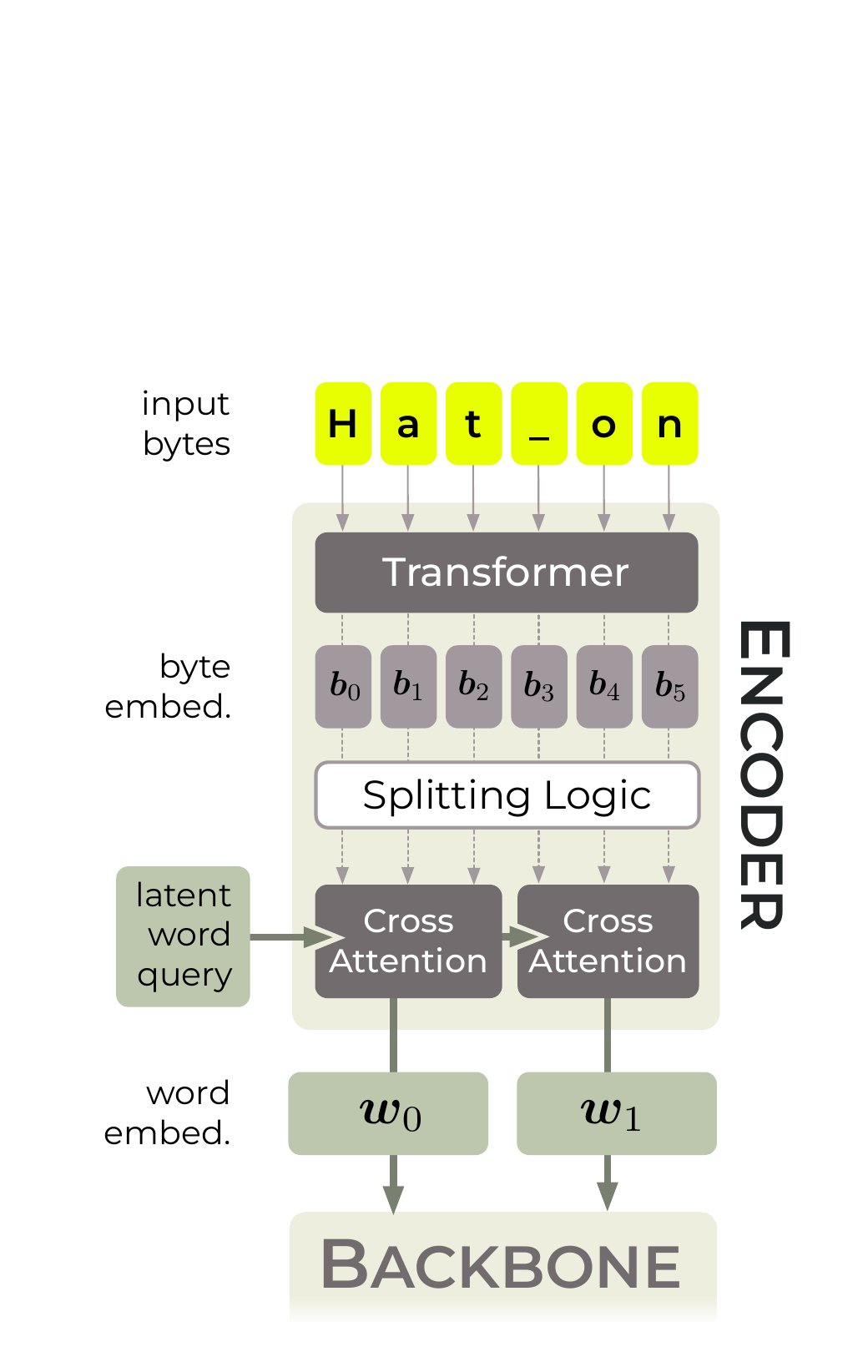}
        \caption{\textbf{Encoder Architecture.} Input text is encoded in UTF8, embedded into the model's vector space and passed through a causal local-sliding-window-attention transformer. The byte embeddings $\bm{b}_i$ are grouped into words according to the split logic. The byte embeddings of each word are aggregated into a single word embedding $\bm{w}_k$ via cross-attention with a learned query vector.}
        \label{fig:encoder-cross-attn}
    \end{subfigure}
    \hfill
    \begin{subfigure}[t]{0.6\textwidth}
        \centering
        \includegraphics[scale=0.25]{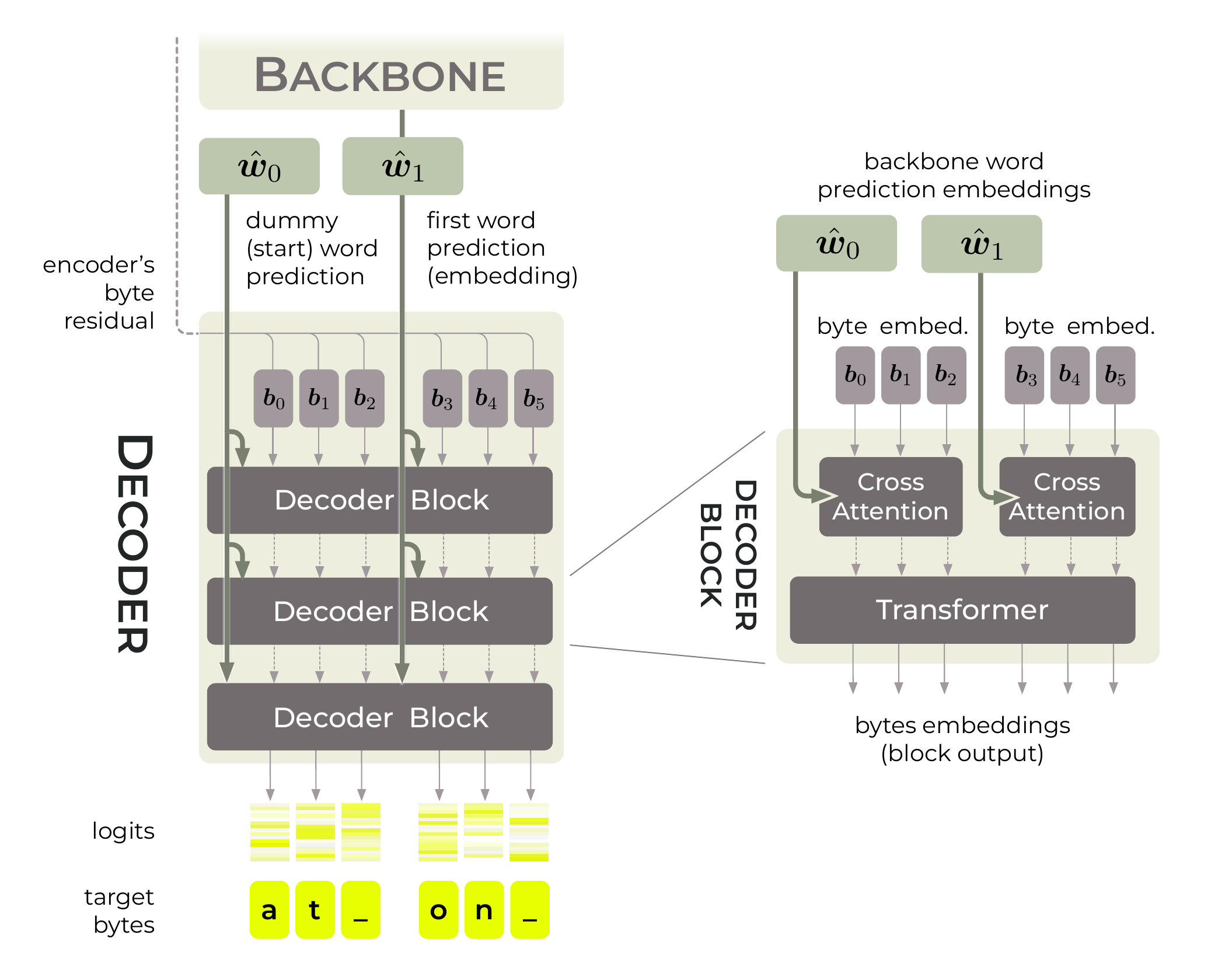}
        \caption{\textbf{Decoder Architecture.} The decoder further processes the byte-level hidden states $\bm{b}_i$ for each word from the encoder with alternating cross-attention and transformer layers. The cross-attention layer uses the shifted next word predictions $\hat{\bm{w}}_{k+1}$ from the backbone as keys and values, and the byte-level hidden states $\bm{b}_i$ as queries. The final hidden states are then passed through a language modeling head to produce byte-level logits.}
        \label{fig:decoder-cross-attn}
    \end{subfigure}
    \caption{Visualization of the encoder and decoder of the HAT model.}
    \label{fig:encoder-decoder-architecture}
\end{figure}

\subsection{Architecture} \label{subsection:architecture}
Our models use a modified HAT architecture \cite{neitemeierhierarchical} consisting of three components: encoder, backbone, and decoder together with connector layers between components (see Figure \ref{fig:full-architecture} and Figure \ref{fig:encoder-decoder-architecture}). Encoder, backbone, and decoder are all instances of dense autoregressive transformers with pre-norm residual blocks in the style of Llama, using a SwiGLU unit as a feed-forward block, with all model parameters active during training and inference. The backbone model uses standard causal attention, while the encoder and decoder use local causal attention with a finite look-back window.

The encoder processes input text as a sequence of UTF-8 bytes and produces a sequence $\bm{b}_i$ of byte embeddings of the same length. This sequence is then split into chunks corresponding to words or other semantic units in the text (we refer to §\ref{subsection:word_splitter} for details). In the encoder-backbone connector layer, for each word, a learned latent vector cross-attends to its corresponding chunk of encoder activations. The resulting sequence of latent vectors $\bm{w}_k$ then serves as input to the backbone. The backbone processes this latent sequence and produces a sequence of next word-level representations $\hat{\bm{w}}_{k+1}$. Finally, the decoder acts on the encoder's byte-level activations $\bm{b}$ and has an LM head that produces next-byte probabilities. To make use of the higher level information stored in the backbone's next word-level embeddings $\hat{\bm{w}}_{k+1}$ during decoding, another cross-attention mechanism is used. Specifically, in each transformer block of the decoder, every byte-level position cross-attends to the backbone's next word-level representations that correspond to the word preceding this byte. The encoder and decoder architectures are detailed in Figure~\ref{fig:encoder-decoder-architecture} and a full architecture overview is shown in Figure~\ref{fig:full-architecture}.

\paragraph{Model details} We present details on the architecture of \HATpretrained and \HATifiedModelSmall in Table \ref{table:7B-8B-architecture-details}, and of \HATifiedModelLarge in Table \ref{table:70B-architecture-details}. We note that \HATifiedModelSmall substantially reduces the number of parameters of the model compared to Llama 3.1 8B and is closer to a 7B than an 8B model: Llama 3.1 8B has a total of 8,030,261,248 parameters, while \HATifiedModelSmall has only 7,192,495,104 parameters. This reduction is proportionally smaller for \HATifiedModelLarge, which has 69,302,847,488 parameters, while Llama 3.1 70B has 71,604,379,648.

\HATifiedModelSmall and Llama 3.1 8B share the same backbone with $32 \times$218,112,000~$=$ 6,979,584,000 parameters, whereas \HATifiedModelLarge and Llama 3.1 70B share a backbone with 68,452,352,000 parameters. In both cases, the Llama 3.1 models have a vocabulary size of 128,256. Llama 3.1 8B has a hidden dimension of 4,096, resulting in a token embedding matrix with 525,336,576 parameters, while Llama 3.1 70B has a hidden dimension of 8,192, resulting in a token embedding matrix with 1,050,673,152 parameters. \HATifiedModelSmall replaces this with an encoder with only 119,291,904 parameters in total. Similarly, the language model head of Llama 3.1 8B with 525,336,576 parameters is replaced by a decoder with a total of 93,619,200 parameters. This reduces the parameter footprint of non-backbone weights from 13\% in Llama 3.1 8B to less than 3\% in the HATified model. We note that the competitive performance of \HATifiedModelSmall is an indicator that the embedding and language model head of Llama 3.1 8B are over-parametrized, presumably since they do not take token similarity into account.

\begin{table}[tb]
\caption{Encoder, backbone, and decoder module configurations of \HATpretrained and \HATifiedModelSmall.}
\centering
\begin{tabular}{lccc}
\toprule
 & \textbf{Encoder} & \textbf{Backbone} & \textbf{Decoder} \\
\midrule
\# of layers                  & 6              & 32             & 4              \\
\# of self-attention heads    & 8              & 32             & 8              \\
Head size                     & 128            & 128            & 128            \\
\# of key-value heads         & 8              & 8              & 8              \\
Hidden size                   & 1,024          & 4,096          & 1,024           \\
Cross-attention hidden size   & 4,096          & --             & 4,096          \\
\# of cross-attention heads   & 32             & --             & 8          \\
MLP expansion factor          & 2.75           & 3.5            & 2.75           \\
MLP type                      & SwiGLU         & SwiGLU         & SwiGLU         \\
Sequence length               & 262,144        & 32,900         & 262,144         \\
Position embeddings           & RoPE base 1e5  & RoPE base 5e5  & RoPE base 1e5  \\
Attention type                & causal + local (768) & causal & causal + local (768) \\
Number of parameters & 119,291,904 & 6,979,584,000 & 93,619,200 \\
\bottomrule
\end{tabular}
\label{table:7B-8B-architecture-details}
\end{table}

\begin{table}[tb]
\caption{Encoder, backbone, and decoder module configurations of \HATifiedModelLarge.}
\centering
\begin{tabular}{lccc}
\toprule
 & \textbf{Encoder} & \textbf{Backbone} & \textbf{Decoder} \\
\midrule
\# of layers              & 6              & 80             & 4              \\
\# of self-attention heads     & 16              & 64             & 16              \\
Head size                     & 128            & 128            & 128            \\
\# of key-value heads     & 16              & 8              & 16              \\
Hidden size                   & 2,048           & 8,192           & 2,048           \\
Cross-attention hidden size   & 8,192           & --             & 2,048          \\
\# of cross-attention heads & 64           & --             & 16          \\
MLP expansion factor          & 2.75           & 3.5            & 2.75           \\
MLP type                      & SwiGLU         & SwiGLU         & SwiGLU         \\
Sequence length               & 98,304         & 12,288          & 98,304         \\
Position embeddings           & RoPE base 1e5  & RoPE base 5e5  & RoPE base 1e5  \\
Attention type                & causal + local (768) & causal & causal + local (768) \\
Number of parameters & 476,610,560 & 68,452,352,000 & 373,884,928 \\
\bottomrule
\end{tabular}
\label{table:70B-architecture-details}
\end{table}

In their architectures, \HATpretrained and \HATifiedModelSmall are almost identical. However, there were two small differences: unlike in the \HATifiedModelSmall, in \HATpretrained we (1) added QK-norm (per head) and (2) used attention logit softcapping at 100 during pre-training (but not during long-context adaptation or post-training, i.e., no inference-relevant change), which we found to be important for training stability.

\subsection{Word splitter}\label{subsection:word_splitter}
To split arbitrary byte sequences, we adopted the guidelines from UAX\#29\footnote{\url{https://unicode.org/reports/tr29/}}, which splits text into words for common Western languages but also produces meaningful semantic units for other common languages (e.g., Chinese, Japanese, Korean). From now on, we refer to these splits as words. We also merged leading whitespace and trailing punctuation into the words to reduce sequence length at the word level. Our splitter also splits camel case like ``FooBar'' into ``Foo'' and ``Bar'', and treats mathematical symbols (as defined in the unicode standard) as separate words. To minimize the computational overhead of this word splitting, we developed an efficient implementation in Rust with Python-bindings, that we make publicly available as hat-splitter\footnote{\url{https://github.com/Aleph-Alpha-Research/hat-splitter}}.

\subsection{Tokenization vs. Splitting}\label{subsection:tokenization}

As noted by the reviewers\footnote{\url{https://openreview.net/forum?id=tU074jg2vS}} of Neitemeier et al.\ \cite{neitemeierhierarchical} -- the basis of our work -- one can interpret our approach as rule-based tokenization. Therefore, and given the potential ambiguity of what `tokenization' as a process means, we give practical definitions of classical tokenization and our alternative, which is a split-based pre-processing approach which we denote as ``tokenizer-free''.
\begin{defn}[General tokenizer]
    Let $X$ be any input space and $\mathcal{V} = \{0,..,N\}$ a finite vocabulary. Then a tokenizer is a map $\tau \colon X \to \mathcal{V}^+$, with $\mathcal{V}^+$ the set of sequences of elements of $\mathcal{V}$ with one or more elements.
\end{defn}
Here, $X$ could be any space of data, for example text or images. We denote by $\mathcal{B} = \{0,..,255 \}$ the byte vocabulary. For language models, we assume that the raw data space is given by $X = \mathcal{B}^L$ for some large but finite value of $L$. A standard tokenizer with a vocabulary of token $\mathcal{V}$, e.g., using byte-pair encoding (BPE) \citep{gage1994new}, can then be naturally interpreted as a map $\mathcal{B}^L \to \mathcal{V}^{\leq L}$, i.e. a sequence of tokens of length at most $L$. We reduce ourselves, without loss of generality, to the case where a tokenizer produces a sequence of tokens of length always less than the original bytes sequence\footnote{The more general case is when a tokenizer creates a bounded number of tokens for each byte, most of the time reducing the sequence length; some tokenizers do create additional tokens such as "Hat" $\rightarrow$ "<capital>hat".}. In contrast we define a splitting rule as follows:
\begin{defn}[Splitting rule]
    For a given sequence $x \in \mathcal{B}^L$ we define the $\mathcal{P}(x)$ as the set of all ordered tuples of non-empty contiguous byte subsequences whose concatenation equals $x$.

    Then, with $\mathcal{B}^{\leq L}$ being the space of sequences of bytes of length at most $L$, a splitting rule is a map $s\colon \mathcal{B}^L \to (\mathcal{B}^{\leq L})^{\leq L}$, such that $s(x) \in \mathcal{P}(x)$ for all $x$.
    
\end{defn}

One could argue that tokenization and splitting are the same conceptually: classical tokenizers such as BPE learn a vocabulary of subwords that the byte sequence is split into. And if we define a vocabulary $\mathcal{V} = \mathcal{B}^{\leq L}$, splitting rules are standard tokenizers. Although technically correct, this does not translate in practice to a useful equivalence, as one would need to create an embedding table with $|\mathcal{B}^{\leq L}| \simeq 256^{L}$ rows\footnote{The output space of possible chunks could in theory be a smaller subset of $\mathcal{B}^{\leq L}$, for example when using a tokenizer as splitting rule. Our word splitter (Section~\ref{subsection:word_splitter}) however can indeed produce every possible chunk.}, which for $L=3$ translates to 16 million entries, two orders of magnitude more than the vocabulary size of SOTA LLMs and computationally intractable.

Additionally, such an interpretation treats all possible byte chunks as categorically different, whereas natural notions of distance (e.g., Jaccard similarity) exist between byte chunks that can be leveraged by a suitable architecture such as ours. Traditional tokenizer-based models first segment text into subwords and then map each subword to an integer token ID. These IDs are used to index a lookup table. Because the IDs are assigned arbitrarily, this representation does not encode any intrinsic similarity between tokens at initialization. For example, the word "car" might be mapped to token ID 7063 and "cars" to token ID 51808; their embeddings are therefore unrelated prior to training. In contrast, architectures such as ours compute embeddings directly from raw byte information rather than from arbitrarily-assigned token IDs. This makes character-level structure available to the model and allows architectures to exploit shared byte patterns. For instance, the overlap between the characters in "car" and "cars", can contribute to more similar representations even before learning has taken place. This allows the model to better generalize to unseen or rare words and capture fine-grained linguistic patterns by design.

\subsection{Infrastructure and Code Optimizations} \label{subsection:code_optimizations}

The HAT architecture differs significantly from standard tokenizer-based language models. Our codebase accounts for this by introducing a number of targeted code optimizations. In tokenizer-based transformer architectures, each transformer layer is associated with the same compute patterns, i.e., model dimensions and sequence lengths stay constant. In contrast, HAT model encoder and decoder layers have far fewer parameters than backbone layers, but require more activation memory since they operate on the byte level. This introduces a trade-off: with an increasing number of bytes in a batch, encoder and decoder layers can require more GPU memory than the backbone due to activations; but they are less compute bound due to smaller hidden dimensions and fewer layers. 

We trained our final model with pipeline parallelism of 2 (with equal number of parameters in each pipe stage) combined with data parallelism and ZeRO-1 sharding \citep{rajbhandari2020zeromemoryoptimizationstraining} (i.e., we only shard optimizer states). Since we train models in mixed precision \citep{micikevicius2018mixedprecisiontraining} with bfloat16 weights and gradients, we keep float32 parameter copies in the optimizer states. We run micro batches of size 1 and utilize gradient accumulation to hide pipe parallelism bubbles. 

We also ran into illegal memory accesses in our cross-attention layers due to non-traditional shapes that are not supported by the flash attention kernel \citep{dao2022flashattentionfastmemoryefficientexact}. In particular, our cross-attention layers work with query projections that contain as many as 80,000 sequences of length 1. This breaks the assumptions of the flash attention kernel, which expects fewer and longer sequences. In the backward pass, flash attention launches parallel threads proportional to the number of sequences, which can end up exceeding the limit of available threads on the GPU in our use case. To mitigate this, we adapted the flash attention kernel to check for number of sequences in the query projection during the backward pass and cap the number of parallel thread launches to avoid illegal memory accesses. This was crucial to enable pre-training with reasonable context lengths ($\approx$ 3,500 words for the backbone). 

HAT also presents a challenge for long context adaptation due to non-constant sequence length during the forward pass (encoder operates on bytes, then backbone on words, then decoder again on bytes). Usually, for long context adaptation, an increase in sequence length leads to more memory consumption for activations, which becomes a bottleneck. Sharding the sequence across multiple GPUs, that is, keeping different tokens on different GPUs, alleviates this bottleneck. This technique is commonly known as context parallelism~\cite{hf2025ultrascale}. In context parallelism, the self-attention operation must communicate between different context-parallel ranks since self-attention requires pairwise interactions between all tokens in a sequence. Recent works such as ring attention~\cite{liu2023ringattentionblockwisetransformers} and striped attention~\cite{brandon2023stripedattentionfasterring} have proposed methods to do this efficiently by overlapping compute with communications. Naïvely running context parallelism via ring attention \citep{liu2023ringattentionblockwisetransformers} requires first calling an \texttt{all\_concat} along the sequence dimension after the encoder layers, resharding the sequence along words, calling \texttt{all\_concat} again after the backbone, and finally resharding along bytes for the decoder. This massively slows down training, leading to suboptimal GPU utilization. Moreover, publicly available ring attention (or other more efficient variants such as striped or zigzag~\footnote{\url{https://github.com/zhuzilin/ring-flash-attention}} attention \citep{brandon2023stripedattentionfasterring}) implementations do not support sliding window attention, which prevents their use in encoder and decoder layers of a HAT model. Due to these complexities, we opted not to employ context parallelism and instead use model parallelism to shard activation memory for long context when necessary.

\section{Pre-training} \label{sec:pre-training}

\subsection{Pre-training data}\label{subsec:pre-training-data}

We trained our models on a filtered subset of diverse corpora of German and English text data including proprietary curated datasets, publicly available high-quality web content, public domain sources, mathematical texts, and programming code, as summarized in Table~\ref{tab:pre-training-data-summary}.

\begin{table}[h]
\caption{Proportions and sources of data used in pre-training.}\label{tab:pre-training-data-summary}
\centering
\begin{tabular}{llr}
\toprule
\textbf{Category} & \textbf{Source} & \textbf{Proportion} \\
\midrule
\multirow{2}{*}{English (70\%)} 
  & Curated web and synthetic data & 63.0\% \\
  & Curated sources (e.g., public domain books) & 7.0\% \\
\midrule
\multirow{2}{*}{German (7\%)} 
  & Curated web and synthetic data & 6.3\% \\
  & Curated sources (e.g., public domain books) & 0.7\% \\
\midrule
\multirow{2}{*}{Mathematics (5\%)} 
  & Mathematical code and proofs & 2.0\% \\
  & Mathematical word problems and equations & 3.0\% \\
\midrule
\multirow{2}{*}{Programming (18\%)} 
  & General programming code & 11.0\% \\
  & High-quality and synthetic Python code & 7.0\% \\
\bottomrule
\end{tabular}
\end{table}

As part of our training recipe, we up-weighted the proportion of English data, which is a common practice in multilingual LLM training due to its broad task coverage and higher availability of quality data, which are essential for effective cross-lingual transfer \citep{ravisankar2025mapenglishrolecrosslingual}. In addition, we included a substantial amount of mathematics and code data, which has become a standard inclusion even when the model is intended only for natural language, as it has been shown to improve performance on a range of non-coding and non-mathematical downstream tasks when included during pre-training \citep{ma2023trainingstagedoescode,aryabumi2024codecodeexploringimpact,kim2024codepretrainingimprovesentity,ruis2025proceduralknowledgepretrainingdrives}. We also augmented the organic data with synthetic data generated by permissively-licensed LLMs.

To ensure pre-training data quality, we applied a range of curation techniques, as suggested by state-of-the-art pre-training data curation methods \citep{nemotronCC,GermanWeb}. These include but are not limited to: 
\begin{itemize}
    \item \textbf{URL filtering}. We used a URL filter developed to filter out fraudulent, harmful, and illegal content from an explicit blocklist, e.g., adult or copyright-infringing websites, or URLs containing words associated with fraudulent, harmful, or adult content.
    \item \textbf{Text extraction}. We extracted natural language texts which were embedded in HTML and other web programming languages using the Resiliparse text extractor  \citep{Resiliparse}.
    \item \textbf{Language identification}. We used a fastText language classifier trained on character n-grams from publicly-available data \cite{fasttext2016} to identify, retain, and sort texts into English and German.
    \item \textbf{Repetition removal}. We applied heuristic methods for detection and removal of documents which contained repetitions on the paragraph, line, word, and character level.
    \item \textbf{Document- and line-level filtering}. We relied on additional document-level heuristics to ensure documents had reasonable numbers and qualities of words, naturalistic symbols-to-words and numbers-to-words ratios, were not predominantly made up of bullet points (to avoid e.g. table of content data, or website menus), and had a sufficient quantity of real words.
    \item \textbf{Deduplication}. We removed duplicate documents via exact and fuzzy deduplication.
    \item \textbf{Model-based filtering}. We trained and used various models to identify the qualities and characteristics of documents to filter out low-quality and less informative data.
\end{itemize}

\subsection{Training Recipe}

We randomly initialized all model parameters of the encoder, decoder, and connector layers. The backbone architecture precisely matches the Llama 3.1 architecture, which allows us to initialize the weights to the pre-trained Llama 3.1 weights for the HATified model variants. We then trained the model on a next-byte-prediction cross-entropy objective on a large and diverse document corpus (see above). Initially, we trained on sequences up to 3,500 words for a total of 134B words. We then continued training on sequences of up to 32,768 words for another 20B words, upweighting longer documents to make use of the extended context. We conducted the training in our Scaling framework\footnote{\url{https://github.com/Aleph-Alpha-Research/scaling}}.

\paragraph{Initial Pre-training}
We used a global batch-size of 512 for \HATifiedModelSmall, and 1,024 for \HATpretrained and \HATifiedModelLarge. Since the sequence length in the first phase of pre-training for all models was 3,500 words, each batch for \HATifiedModelSmall had 1,792,000 words. Each batch of \HATpretrained and \HATifiedModelLarge had 3,584,000 words.

We employed a warmup-stable-decay learning rate scheduler \cite{hu2024minicpm,wen2025understanding} for all models. We initially trained \HATpretrained for 1,000,000 steps, \HATifiedModelSmall for 75,000 steps, and \HATifiedModelLarge for 35,000 steps. During the learning rate decay phase, we maintained the same data sources from Table~\ref{tab:pre-training-data-summary}, but upsampled German and code data. For encoder and decoder, we employed a linear warmup of 500 steps, a stable phase with learning rate 3e-4, and a final decay phase -- of 50,000 steps for \HATpretrained, of 10,000 steps for \HATifiedModelSmall, and of 7,000 steps for \HATifiedModelLarge. For \HATpretrained, we employed weight decay of 0.05 for all parameters except for the embedding and normalization parameters.

For \HATifiedModelSmall and \HATifiedModelLarge, we kept the backbone frozen for the first 2,000 steps, but otherwise followed the same schedule with the stable phase (shortened by 2,000 steps accordingly and learning rate decreased to 3e-5). We used a delayed training and reduced learning rate for the backbone because in the HATification case it is the component which is not initialized randomly -- the encoder/decoder requires some head-start for learning to ``utilize'' the pre-trained backbone before we can start adapting the backbone.\footnote{The training dynamics show a distinct ``bump'' in performance when the encoder/decoder learn to utilize the pre-trained backbone, indicating the transition from a pure byte-level model to an actual hierarchical behavior. This happens usually within the first 2,000 steps.} We used the training data mix for this phase as described in §\ref{subsec:pre-training-data}.

\paragraph{Long-context adaptation}
For adapting the models to longer word context lengths, we kept most parameters frozen and only adapted query and key projections of cross-attention layers in encoder and decoder and of self-attention layers in the backbone. The long-context length for \HATpretrained was 32,900 words, for \HATifiedModelSmall was 32,768 words, and for \HATifiedModelLarge was 16,000 words. We once again adopted a warmup-stable-decay schedule for this adaptation with 500 steps of linear warm-up, then a stable learning rate of 3e-6, and 1,000 steps of decay to 0. In this phase, we use a global batch size of 128 and up-weighted long-sequence subsets of all data sources while maintaining a similar mix.

\section{Post-training} \label{sec:post-training}

We optimized our models for instruction-following using a standard post-training pipeline. First, we applied supervised fine-tuning (SFT) to train the model on both single-turn and multi-turn (chat) instruction-following tasks. Next, we aligned our model for helpfulness and to answer safely using DPO.

\subsection{Supervised Fine-tuning}\label{subsec:sft}

The data we used for instruction fine-tuning is based on a mixture of publicly-available user prompts and model completions. The data mixture consisted of roughly 2M samples from diverse datasets including but not limited to: human feedback focused on helpful and harmless responses; a small curated set for specific response patterns; safety and robustness subsets for appropriate boundaries; specialized datasets covering mathematics, programming, and logical inference; collaborative conversational data; formal mathematics with advanced problems; multilingual conversation prompts; and tabular data reasoning for structured information.

For datasets comprising answers from older model generations, we synthesized responses to English queries using Qwen 2.5-32B and Qwen 2.5-72B \cite{qwen2.5}. We chose the Qwen 2.5 models for data synthesis, because they offer strong performance across a variety of tasks \citep{qwen2025qwen25technicalreport} at relatively low resource requirements. Furthermore, we focused on improving both the coverage and quality of German SFT data. For this, we (i) translated English prompts in our datasets using Mistral-Nemo-Instruct-2407 \citep{mistralneMo_2024}, (ii) generated corresponding answers using Mistral-Small-3.1-Instruct \citep{mistral_small_3_1_2025}, and (iii) performed quality filtering on the prompt-answer pairs using an LLM judge based on Llama-3.3-70B-Instruct \citep{llama3modelcard}. 
We found that translating only the prompt first and then using it to generate a response directly in German significantly outperforms the alternative approach of translating both the prompt and the answer.

We also found that quality filtering on the generated answers is best done by rejecting all but the highest quality samples, whereas it can be beneficial to trade off data quantity for quality in the case of the translated prompts. We chose Mistral-Nemo-Instruct-2407 and Mistral-Small-3.1-Instruct due to strong performance-to-size characteristics on internal English-to-German translation and German Q\&A benchmarks, respectively. We chose Llama-3.3-70B-Instruct for filtering due to its high agreement with human raters. Lastly, we supplemented the synthetic data with proprietary human-generated SFT data.

We experimented with various curriculum training methods, but found no significant advantage -- for example, in training more general tasks first and complex, specific instructions last. As a result, our final dataset randomly intersperses samples across datasets. We used a packed training procedure, wherein multiple shorter sequences are concatenated into a single context window. Consequently, this means the number of sequences per batch can vary depending on how well the packing fills the context length. Based on our data distribution, this corresponds to an approximate batch size of $256$ sequences. The default learning rate was 3e-6, although for the HATified models we used a learning rate of 1.5e-6 for the backbone.

\subsection{Direct Preference Optimization}\label{subsec:dpo}

For alignment training, we used the length-normalized version of DPO as suggested by Lambert et al.~\cite{lambert2025tulu3}. As for SFT, our alignment dataset of prompts and completions originates from diverse domains. We filtered out any preference samples that contained empty strings or where the chosen and rejected completions were the same string. We used a learning rate of 1e-6 and, as for SFT, an approximate batch-size of $\sim 256$ (i.e., with sequence packing).

\section{Inference}\label{sec:inference}

Our initial release included a simplified HuggingFace inference implementation for single requests to demonstrate our architecture. For production deployment with efficient batched serving, we selected vLLM\footnote{ \url{https://github.com/vllm-project/vllm}}~\cite{kwon2023efficient} as our serving framework. The integration revealed several fundamental architectural challenges unique to hierarchical models that required careful adaptation of vLLM's components, with a key constraint being to structure our implementation to minimize modifications to the core vLLM codebase.

The hierarchical nature of HAT---and similar architectures like the Byte Latent Transformer (BLT) \citep{pagnoni2024bytelatenttransformerpatches}---introduces fundamental batching challenges that require changes to the scheduling strategy~\cite{yu2022orca}. Unlike traditional autoregressive models that generate one token per step, HAT generates a variable number of bytes before reaching word boundaries, creating synchronization complexities. The system must balance between maximizing backbone utilization (by waiting for all sequences in a batch to reach word boundaries) and maintaining reasonable latency (by processing a fixed number of bytes). This variability also means that sequences within the same batch may require different computational patterns at any given step (some continuing byte-level generation while others are ready for word-level processing), necessitating careful orchestration to maintain efficiency. For example, prefills, chunked-prefills, and decodes that have just crossed a word boundary must traverse the full encoder-backbone-decoder pipeline. However, decodes that are in the middle of a word, and therefore do not require another consultation with the backbone, can keep cycling through the lightweight encoder-decoder loop until they reach the next word boundary.

Additionally, the dual-sequence architecture of HAT required maintaining two distinct sequence representations simultaneously: one for tracking byte-level processing through the encoder-decoder components and another for word-level processing through the backbone model. A significant departure from traditional LLM serving lies in the corresponding KV cache management, where HAT requires maintaining separate caches for both sequences. This dual-cache system must ensure coordinated memory allocation, as the mapping between sequences is non-uniform (a single word may comprise many bytes) and a sequence can only proceed if sufficient memory exists for both its byte and word representations. The challenge is compounded by asymmetric memory requirements: word-level cache blocks consume significantly more memory mainly due to global attention requiring linearly increasing memory, while byte-level cache blocks are capped by the sliding window~\cite{beltagy2020longformer}.

Due to these challenges, the current implementation in vLLM exhibits lower throughput compared to a FLOP-matched tokenizer-based transformer in the batched setting. Nevertheless, to the best of our knowledge, it is the fastest inference implementation available for any hierarchical architecture of this kind. We are actively working to optimize the implementation further and are also designing future iterations of the architecture with greater inference efficiency in mind.

\section{Model Performance}\label{sec:performance}

We evaluate our pre-trained and post-trained (SFT, DPO) models with different subsets of benchmark tasks. To accomplish this consistently, we chose not to use a combination of several existing evaluation suites but rather develop and use a unified framework, which we release under Apache 2.0\footnote{\url{https://github.com/Aleph-Alpha-Research/eval-framework/}}.

Our evaluation framework aligns with the EleutherAI's LM Evaluation Harness \cite{eval-harness} benchmarks implementations whenever possible. However, we permitted minor prompt modifications when such changes improved applicability to practical, model-agnostic scenarios. These adjustments primarily involve whitespaces, newlines, and translated cue words. For example, in non-English benchmarks, we often encountered English cue words such as “Question” and “Answer” where language-specific terms like the German cue words “Frage” and “Antwort” seem more appropriate. For full details of our implementations of each benchmark, please consult our code.

\subsection{Towards consistent evaluations}

We invested significant effort to unify a large set of evaluation tasks by enforcing the same prompt (especially formatting) conventions, inference settings across back-ends, completion parsing algorithms, metric implementations, and error handling. Removing these sources of variability enabled us to better understand the model performance across benchmarks and, in effect, iterate faster and have greater confidence in our evaluations.

While we draw on established prompts, commonly used formats, and standard techniques, we reserve the flexibility to adapt them when necessary to better reflect realistic use cases and ensure fair and meaningful evaluations. As a result, we may not reproduce the exact numbers reported for some models on certain benchmarks. However, we believe that our comparisons remain fair, as our evaluation framework ensures that all models are tested under consistent and controlled conditions. We believe this is important not just for transparency, but also so others may reproduce our results using our publicly-available checkpoints and accompanying evaluation framework.

\paragraph{Whitespace in prompts:} LLMs are often overly sensitive to seemingly minor details in prompt formatting. One notable example we have observed (as have others \citep{chang-etal-2025-mind}) relates to the placement of whitespace in prompt or completion sequences. In multiple-choice benchmarks, such as MMLU, the task is typically presented as a set of token sequences that represent each answer option. The log-likelihood of each sequence is calculated and compared, and the sequence with the highest log-likelihood is selected as the model's predicted answer. Some benchmarks guide models in the desired direction by ending their prompt with an open-ended assistant's message, a technique called assistant-prefilling or answer-cueing. However, there is a difference in token log-likelihood if the cue ends with a space ("Answer:~" followed with tested completions without space prefixes) or not ("Answer:" followed with tested completions with space prefixes). We observed that for models trained with space prefix tokenizers, the former case leads to very low log-likelihoods (indicating out of distribution tokens) and less reliable comparisons of those. We adjusted evaluations so that they would not have a trailing space in the prompt, and would have, in the case of log-likelihood tasks, a prefix space in the possible continuations.

\subsection{Performance and Compression}

\paragraph{Performance:} Our T-Free models deliver performance on par with state-of-the-art open-weight memory-equivalent models in both English and German. For evaluation purposes, we compare our tokenizer-free base models to Llama 3.1 8B Base, our SFT model to Tülu 3.1 8B SFT \cite{lambert2025tulu3}, and our DPO model to Llama 3.1 8B Instruct and Tülu 3.1 8B \cite{lambert2025tulu3}. The respective benchmarks and results can be found in the tables below. Note that although we used code and mathematics data in our training corpus, our model's architecture has not been optimized for code generation and mathematical reasoning and is therefore not evaluated extensively on those benchmarks.

\paragraph{Compression:} Our T-Free approach results in improved efficiency, particularly in inference overhead, measured in terms of number of words processed across all languages and domains. We define efficiency as tokenizer fertility, or bytes per sequence position in the backbone, with higher value indicating better performance. In-production latency and throughput are currently beyond the scope of research-centric evaluations and will be addressed in the future. At present, our evaluation framework automatically measures bytes per sequence position across datasets, allowing us to derive efficiency scores and analyze variations across different dataset distributions. In Appendix \ref{app:evals}, we provide more comprehensive descriptions of the benchmarks and metrics used to evaluate our models.

\subsection{Evaluation Results}

\paragraph{Few-shot prompting:} In some evaluations, we have used few-shot settings, closely aligned to literature of models we compare against and what is commonly used, e.g., for submitting to OpenLLMLeaderboard \cite{open-llm-leaderboard-v2} or the Llama 3 tech report \cite{grattafiori2024llama3herdmodels}. The number of few-shot examples provided during evaluation is detailed in the tables below. In each row, bold numbers show the highest score and any score within 1\% of the highest score.

\subsubsection{Pre-trained Model}

For the sake of conciseness, in the figures and table below, "HAT 7B Base" or "T-Free" refers to \HATpretrained-Base, our trained from scratch model, while "HATified 7B Base" or "HATified" is \HATifiedModelSmall-Base, and "Llama" or "Llama 8B Base" is Llama-3.1-8B-Base.

\begin{table}[h!]
\caption{Categories and corresponding benchmark tasks used in pre-trained model evaluations.}
\centering
\begin{tabular}{l p{0.75\textwidth}}
\toprule
\textbf{Group} & \textbf{Benchmarks} \\
\midrule
English Knowledge & MMLU \cite{hendrycks2021measuringmassivemultitasklanguage}, Full Text MMLU, MMLU-Pro \cite{wang2024mmluprorobustchallengingmultitask}, Graduate-Level Google-Proof Q\&A (GPQA) \cite{rein2023gpqagraduatelevelgoogleproofqa}, BIG-Bench Hard (BBH) \cite{suzgun2022challengingbigbenchtaskschainofthought}, OpenBookQA \cite{mihaylov2018suitarmorconductelectricity}, TriviaQA \cite{joshi2017triviaqalargescaledistantly}, TruthfulQA \cite{lin2022truthfulqameasuringmodelsmimic} \\
English Reasoning & AI2 Reasoning Challenge (ARC) \cite{allenai:arc}, WinoGrande \cite{sakaguchi2019winograndeadversarialwinogradschema}, HellaSwag \cite{zellers2019hellaswagmachinereallyfinish} \\
German & Multilingual Massive Multitask Language (MMMLU) \cite{hendrycks2021measuringmassivemultitasklanguage}, LAnguage Modeling Broadened to
Account for Discourse Aspects (LAMBADA) \cite{paperno2016lambadadatasetwordprediction}, German ARC (Easy \& Challenge), German Wino-X, German HellaSwag, German TruthfulQA, German GSM8K, WMT16 \\
Math & Grade School Math 8K (GSM8K) \cite{cobbe2021trainingverifierssolvemath} \\
Safety & WinoGender \cite{rudinger-EtAl:2018:N18} \\
\bottomrule
\end{tabular}
\label{tab:pre-trained-model-benchmark-by-group}
\end{table}

\begin{figure}[tb]
    \centering
    \includegraphics[width=\textwidth]{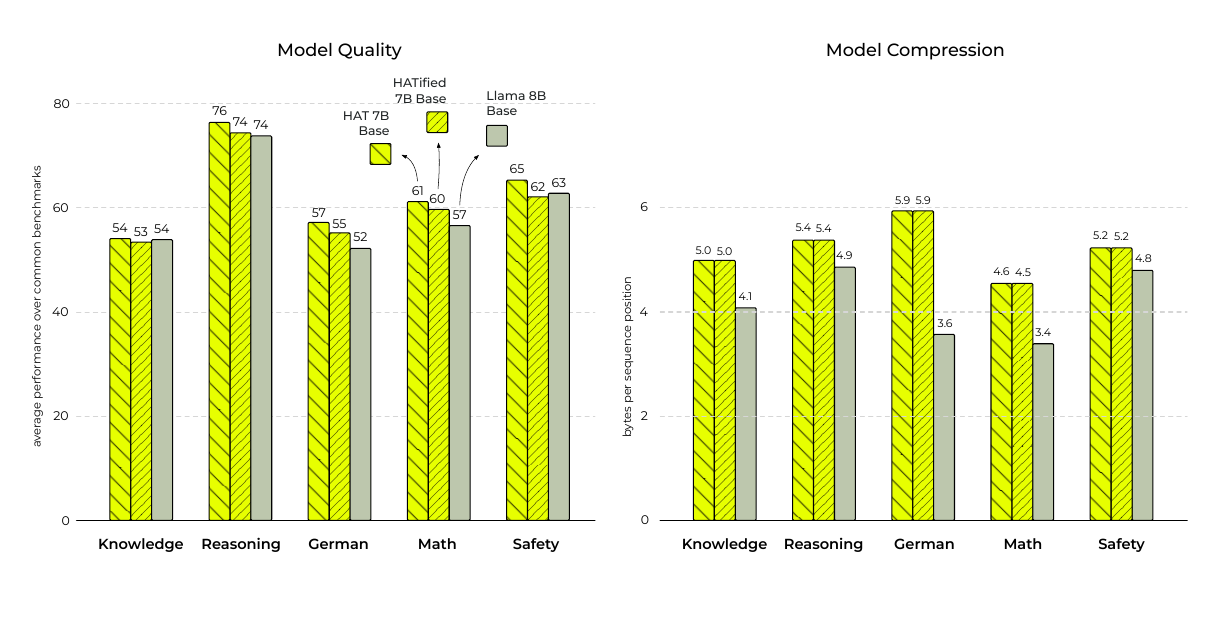}
    \caption{Model Quality and Compression for our pre-trained T-Free model in comparison with Llama-3.1-8B.}
    \label{fig:results-pre-trained}
\end{figure}


\begin{table}[H]
\caption{\textbf{Knowledge}: Pre-training evaluation benchmarks and compression.}
\centering
\resizebox{\textwidth}{!}{
\begin{tabular}{@{}llrrrrrr@{}}
\toprule
\multicolumn{2}{c}{} & \multicolumn{3}{c}{\textbf{Results}} & \multicolumn{3}{c}{\textbf{Compression}} \\
\cmidrule(lr){3-5} \cmidrule(lr){6-8}
\textbf{Task} & \textbf{Metric} & \textbf{T-Free} & \textbf{HATified} & \textbf{Llama} & \textbf{T-Free} & \textbf{HATified} & \textbf{Llama} \\
\midrule
MMLU 5-shot & {norm. log. acc.} & \textbf{0.665} & 0.657 & \textbf{0.670} & \textbf{5.18} & \textbf{5.18} & 4.28 \\
Full Text MMLU 5-shot & {norm. log. acc.} & \textbf{0.649} & 0.638 & 0.623 & \textbf{5.31} & \textbf{5.31} & 4.56 \\
MMLU Pro 5-shot & {norm. log. acc.} & \textbf{0.386} & 0.369 & 0.369 & \textbf{4.73} & \textbf{4.73} & 3.73 \\
GPQA 0-shot & {log. acc.} & 0.273 & \textbf{0.308} & 0.301 & \textbf{4.93} & \textbf{4.93} & 3.52 \\
BBH 3-shot & {norm. log. acc.} & \textbf{0.476} & \textbf{0.472} & \textbf{0.472} & \textbf{4.67} & \textbf{4.67} & 3.79 \\
OpenBookQA 10-shot & {norm. log. acc.} & \textbf{0.894} & 0.868 & 0.846 & \textbf{4.85} & \textbf{4.85} & 4.35 \\
TriviaQA 5-shot & {comp. acc.} & 0.647 & 0.623 & \textbf{0.695} & \textbf{5.37} & \textbf{5.37} & 4.24 \\
TruthfulQA 6-shot & {norm. prob. mass} & \textbf{0.336} & \textbf{0.336} & \textbf{0.337} & \textbf{4.91} & \textbf{4.91} & 4.18 \\
\bottomrule
\end{tabular}
}
\label{tab:pre-training-evaluation-results-knowledge}
\end{table}

\begin{table}[H]
\caption{\textbf{Reasoning}: Pre-training evaluation benchmarks and compression.}
\centering
\resizebox{\textwidth}{!}{
\begin{tabular}{@{}llrrrrrr@{}}
\toprule
\multicolumn{2}{c}{} & \multicolumn{3}{c}{\textbf{Results}} & \multicolumn{3}{c}{\textbf{Compression}} \\
\cmidrule(lr){3-5} \cmidrule(lr){6-8}
\textbf{Task} & \textbf{Metric} & \textbf{T-Free} & \textbf{HATified} & \textbf{Llama} & \textbf{T-Free} & \textbf{HATified} & \textbf{Llama} \\
\midrule
ARC Easy 25-shot & {norm. log. acc.} & \textbf{0.875} & \textbf{0.871} & 0.858 & \textbf{5.53} & \textbf{5.53} & 4.94 \\
ARC Challenge 25-shot & {norm. log. acc.} & \textbf{0.635} & 0.621 & 0.580 & \textbf{5.51} & \textbf{5.51} & 4.92 \\
Winogrande 5-shot & {norm. log. acc.} & \textbf{0.721} & 0.691 & 0.695 & \textbf{5.16} & \textbf{5.16} & 4.91 \\
HellaSwag 10-shot & {norm. log. acc.} & \textbf{0.826} & 0.793 & 0.818 & \textbf{5.34} & \textbf{5.34} & 4.66 \\
\bottomrule
\end{tabular}
}
\label{tab:pre-training-evaluation-results-reasoning}
\end{table}

\begin{table}[H]
\caption{\textbf{German}: Pre-training evaluation benchmarks and compression.}
\centering
\resizebox{\textwidth}{!}{
\begin{tabular}{@{}llrrrrrr@{}}
\toprule
\multicolumn{2}{c}{} & \multicolumn{3}{c}{\textbf{Results}} & \multicolumn{3}{c}{\textbf{Compression}} \\
\cmidrule(lr){3-5} \cmidrule(lr){6-8}
\textbf{Task} & \textbf{Metric} & \textbf{T-Free} & \textbf{HATified} & \textbf{Llama} & \textbf{T-Free} & \textbf{HATified} & \textbf{Llama} \\
\midrule
MMMLU 5-shot & {norm. log. acc.} & \textbf{0.618} & 0.588 & 0.577 & \textbf{6.03} & \textbf{6.03} & 3.33 \\
ARC Easy DE 25-shot & {norm. log. acc.} & \textbf{0.801} & 0.779 & 0.715 & \textbf{6.60} & \textbf{6.60} & 3.68 \\
ARC Challenge DE 25-shot & {norm. log. acc.} & \textbf{0.591} & 0.537 & 0.474 & \textbf{6.57} & \textbf{6.57} & 3.68 \\
Wino-X DE 5-shot & {norm. log. acc.} & \textbf{0.803} & 0.789 & 0.765 & \textbf{5.63} & \textbf{5.63} & 3.67 \\
HellaSwag DE 10-shot & {norm. log. acc.} & \textbf{0.687} & 0.657 & 0.616 & \textbf{6.50} & \textbf{6.50} & 3.67 \\
TruthfulQA DE 6-shot & {norm. prob. mass} & \textbf{0.340} & \textbf{0.341} & \textbf{0.342} & \textbf{5.91} & \textbf{5.91} & 3.39 \\
Lambada 5-shot & {comp. acc.} & \textbf{0.471} & 0.453 & 0.451 & \textbf{5.78} & \textbf{5.78} & 3.56 \\
GSM8K DE 8-shot & {comp. acc.} & \textbf{0.475} & 0.441 & 0.415 & \textbf{4.38} & \textbf{4.37} & 2.94 \\
WMT16 3-shot & {linewise BLEU} & 36.416 & \textbf{38.033} & 34.812 & \textbf{6.02} & \textbf{6.02} & 4.21 \\
\bottomrule
\end{tabular}
}
\label{tab:pre-training-evaluation-results-german}
\end{table}

\begin{table}[H]
\caption{\textbf{Math \& Safety}: Pre-training evaluation benchmarks and compression.}
\centering
\resizebox{\textwidth}{!}{
\begin{tabular}{@{}llrrrrrr@{}}
\toprule
\multicolumn{2}{c}{} & \multicolumn{3}{c}{\textbf{Results}} & \multicolumn{3}{c}{\textbf{Compression}} \\
\cmidrule(lr){3-5} \cmidrule(lr){6-8}
\textbf{Task} & \textbf{Metric} & \textbf{T-Free} & \textbf{HATified} & \textbf{Llama} & \textbf{T-Free} & \textbf{HATified} & \textbf{Llama} \\
\midrule
GSM8K 8-shot & {comp. acc.} & \textbf{0.612} & 0.597 & 0.566 & \textbf{4.55} & \textbf{4.55} & 3.39 \\
Winogender 5-shot & {norm. log. acc.} & \textbf{0.653} & 0.621 & 0.628 & \textbf{5.23} & \textbf{5.23} & 4.80 \\
\bottomrule
\end{tabular}
}
\label{tab:pre-training-evaluation-results-math-safety}
\end{table}

\FloatBarrier
\subsubsection{Post-trained \HATifiedModelSmall: SFT}

As with pre-trained model evaluation, the same benchmark groups are used in post-trained SFT model evaluations, with one additional group, Instruction Following. Furthermore, some groups contain additional benchmarks. In this section, "HATified" is \HATifiedModelSmall-SFT, and "Tülu" is Llama-3.1-Tulu-3-8B-SFT.

\begin{table}[ht!]
\caption{Categories and corresponding benchmark tasks used in post-trained SFT and DPO model evaluations.}
\centering
\begin{tabular}{l p{0.75\textwidth}}
\toprule
\textbf{Group} & \textbf{Benchmarks} \\
\midrule
English Knowledge & Same as pre-training evaluations. \\
English Reasoning & Same as pre-training evaluations. \\
German & Same as pre-training evaluations, as well as the instruct version of WMT16. \\
Instruction Following & Alpaca Eval \cite{dubois2025lengthcontrolledalpacaevalsimpleway} \\
Long Context & QuALITY \cite{pang2022qualityquestionansweringlong}, Ada-LEval \cite{wang2024adalevalevaluatinglongcontextllms} \\
Math & Same as pre-trained evaluations. \\
Safety & Same as pre-trained evaluations. \\
\bottomrule
\end{tabular}
\label{tab:SFT-trained-model-benchmark-by-group}
\end{table}

\begin{figure}[tb]
    \centering
    \includegraphics[width=0.75\textwidth]{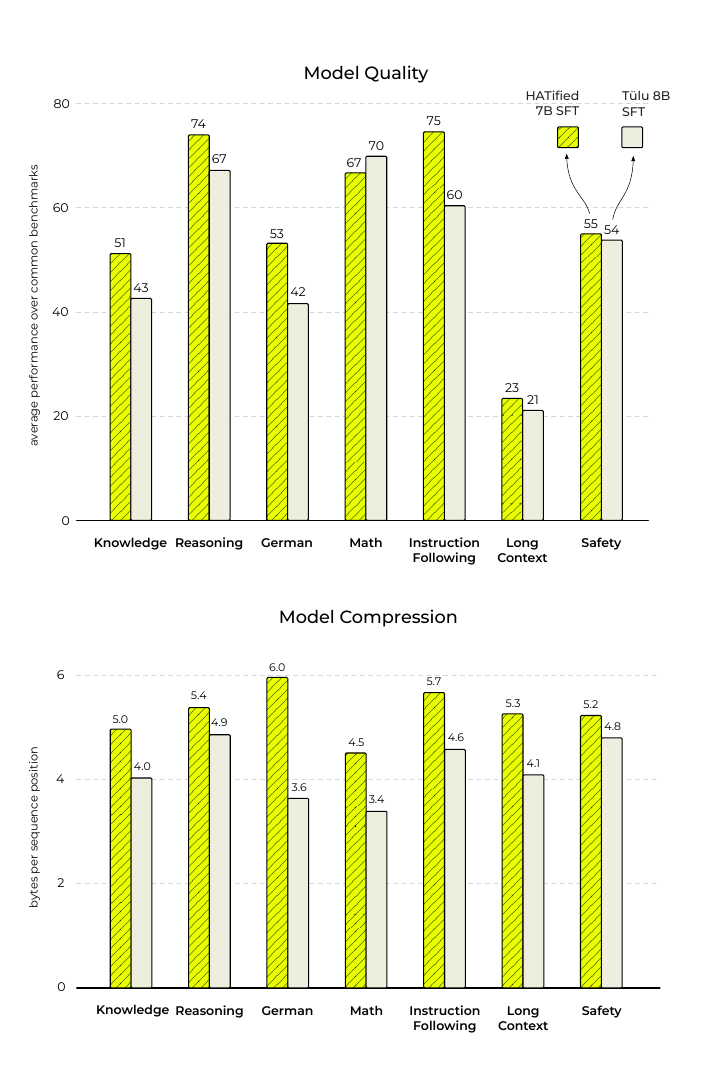}
    \caption{Model Quality and Compression for our SFT T-Free model in comparison with Llama-3.1-Tulu-3-8B-SFT.}
    \label{fig:results-sft}
\end{figure}


\begin{table}[H]
\caption{\textbf{Knowledge}: SFT evaluation benchmarks and compression.}
\centering
\resizebox{\textwidth}{!}{
\begin{tabular}{@{}llrrrr@{}}
\toprule
\multicolumn{2}{c}{} & \multicolumn{2}{c}{\textbf{Results}} & \multicolumn{2}{c}{\textbf{Compression}} \\
\cmidrule(lr){3-4} \cmidrule(lr){5-6}
\textbf{Task} & \textbf{Metric} & \textbf{HATified} & \textbf{Tülu} & \textbf{HATified} & \textbf{Tülu} \\
\midrule
MMLU 5-shot & {norm. log. acc.} & \textbf{0.655} & 0.576 & \textbf{5.18} & 4.28 \\
Full Text MMLU 5-shot & {norm. log. acc.} & \textbf{0.652} & 0.598 & \textbf{5.31} & 4.56 \\
MMLU Pro 5-shot & {norm. log. acc.} & \textbf{0.378} & 0.306 & \textbf{4.73} & 3.73 \\
GPQA 0-shot & {log. acc.} & \textbf{0.294} & 0.277 & \textbf{4.93} & 3.52 \\
BBH 3-shot & {norm. log. acc.} & \textbf{0.493} & 0.462 & \textbf{4.67} & 3.79 \\
OpenBookQA 10-shot & {norm. log. acc.} & \textbf{0.696} & 0.654 & \textbf{4.85} & 4.35 \\
TriviaQA 5-shot & {comp. acc.} & \textbf{0.585} & 0.200 & \textbf{5.36} & 3.93 \\
TruthfulQA 6-shot & {norm. prob. mass} & \textbf{0.346} & 0.338 & \textbf{4.91} & 4.18 \\
\bottomrule
\end{tabular}
}
\label{tab:sft-evaluation-results-knowledge}
\end{table}

\begin{table}[H]
\caption{\textbf{Reasoning}: SFT evaluation benchmarks and compression.}
\centering
\resizebox{\textwidth}{!}{
\begin{tabular}{@{}llrrrr@{}}
\toprule
\multicolumn{2}{c}{} & \multicolumn{2}{c}{\textbf{Results}} & \multicolumn{2}{c}{\textbf{Compression}} \\
\cmidrule(lr){3-4} \cmidrule(lr){5-6}
\textbf{Task} & \textbf{Metric} & \textbf{HATified} & \textbf{Tülu} & \textbf{HATified} & \textbf{Tülu} \\
\midrule
ARC Easy 25-shot & {norm. log. acc.} & \textbf{0.889} & 0.799 & \textbf{5.53} & 4.94 \\
ARC Challenge 25-shot & {norm. log. acc.} & \textbf{0.646} & 0.515 & \textbf{5.51} & 4.92 \\
Winogrande 5-shot & {norm. log. acc.} & \textbf{0.677} & 0.602 & \textbf{5.16} & 4.91 \\
HellaSwag 10-shot & {norm. log. acc.} & 0.747 & \textbf{0.773} & \textbf{5.34} & 4.66 \\
\bottomrule
\end{tabular}
}
\label{tab:sft-evaluation-results-reasoning}
\end{table}

\begin{table}[H]
\caption{\textbf{German}: SFT evaluation benchmarks and compression.}
\centering
\resizebox{\textwidth}{!}{
\begin{tabular}{@{}llrrrr@{}}
\toprule
\multicolumn{2}{c}{} & \multicolumn{2}{c}{\textbf{Results}} & \multicolumn{2}{c}{\textbf{Compression}} \\
\cmidrule(lr){3-4} \cmidrule(lr){5-6}
\textbf{Task} & \textbf{Metric} & \textbf{HATified} & \textbf{Tülu} & \textbf{HATified} & \textbf{Tülu} \\
\midrule
MMMLU 5-shot & {norm. log. acc.} & \textbf{0.597} & 0.468 & \textbf{6.03} & 3.33 \\
ARC Easy DE 25-shot & {norm. log. acc.} & \textbf{0.800} & 0.535 & \textbf{6.60} & 3.68 \\
ARC Challenge DE 25-shot & {norm. log. acc.} & \textbf{0.572} & 0.336 & \textbf{6.57} & 3.68 \\
Wino-X DE 5-shot & {norm. log. acc.} & \textbf{0.763} & 0.657 & \textbf{5.63} & 3.67 \\
HellaSwag DE 10-shot & {norm. log. acc.} & \textbf{0.596} & 0.535 & \textbf{6.50} & 3.67 \\
TruthfulQA DE 6-shot & {norm. prob. mass} & \textbf{0.348} & 0.339 & \textbf{5.91} & 3.39 \\
Lambada 5-shot & {comp. acc.} & \textbf{0.368} & 0.178 & \textbf{5.79} & 3.55 \\
GSM8K DE 8-shot & {comp. acc.} & \textbf{0.550} & 0.477 & \textbf{4.41} & 2.93 \\
WMT16 3-shot & {linewise BLEU} & \textbf{35.810} & 31.414 & \textbf{6.02} & 4.20 \\
WMT16 Instruct 3-shot & {linewise BLEU} & \textbf{36.441} & 32.326 & \textbf{6.11} & 4.30 \\
\bottomrule
\end{tabular}
}
\label{tab:sft-evaluation-results-german}
\end{table}

\begin{table}[H]
\caption{\textbf{Instruction Following}: SFT evaluation benchmarks and compression.}
\centering
\begin{tabular}{@{}llrrrr@{}}
\toprule
\multicolumn{2}{c}{} & \multicolumn{2}{c}{\textbf{Results}} & \multicolumn{2}{c}{\textbf{Compression}} \\
\cmidrule(lr){3-4} \cmidrule(lr){5-6}
\textbf{Task} & \textbf{Metric} & \textbf{HATified} & \textbf{Tülu} & \textbf{HATified} & \textbf{Tülu} \\
\midrule
Alpaca Eval 0-shot & {CS} & \textbf{0.345} & 0.065 & \textbf{5.67} & 4.58 \\
Alpaca Eval 0-shot & {IF} & \textbf{0.903} & 0.846 & \textbf{5.67} & 4.58 \\
Alpaca Eval 0-shot & {LC} & \textbf{0.989} & 0.901 & \textbf{5.67} & 4.58 \\
\bottomrule
\end{tabular}
\label{tab:sft-evaluation-results-instruction-following}
\end{table}

\begin{table}[H]
\caption{\textbf{Long Context}: SFT evaluation benchmarks and compression.}
\centering
\resizebox{\textwidth}{!}{
\begin{tabular}{@{}llrrrr@{}}
\toprule
\multicolumn{2}{c}{} & \multicolumn{2}{c}{\textbf{Results}} & \multicolumn{2}{c}{\textbf{Compression}} \\
\cmidrule(lr){3-4} \cmidrule(lr){5-6}
\textbf{Task} & \textbf{Metric} & \textbf{HATified} & \textbf{Tülu} & \textbf{HATified} & \textbf{Tülu} \\
\midrule
QuALITY 0-shot & {log. acc.} & \textbf{0.389} & 0.346 & \textbf{4.85} & 4.28 \\
Ada-LEval TextSort Choices 0-shot & {log. acc.} & 0.261 & \textbf{0.288} & \textbf{5.19} & 4.04 \\
Ada-LEval TextSort 0-shot & {comp. acc.} & \textbf{0.052} & 0.000 & \textbf{5.20} & 4.06 \\
\bottomrule
\end{tabular}
}
\label{tab:sft-evaluation-results-long-context}
\end{table}

\begin{table}[H]
\caption{\textbf{Math \& Safety}: SFT evaluation benchmarks and compression.}
\centering
\begin{tabular}{@{}llrrrr@{}}
\toprule
\multicolumn{2}{c}{} & \multicolumn{2}{c}{\textbf{Results}} & \multicolumn{2}{c}{\textbf{Compression}} \\
\cmidrule(lr){3-4} \cmidrule(lr){5-6}
\textbf{Task} & \textbf{Metric} & \textbf{HATified} & \textbf{Tülu} & \textbf{HATified} & \textbf{Tülu} \\
\midrule
GSM8K 8-shot & {comp. acc.} & 0.667 & \textbf{0.699} & \textbf{4.51} & 3.39 \\
Winogender 5-shot & {norm. log. acc.} & \textbf{0.550} & 0.537 & \textbf{5.23} & 4.80 \\
\bottomrule
\end{tabular}
\label{tab:sft-evaluation-results-math-safety}
\end{table}

\begin{table}[H]
\caption{\textbf{SFT MTBench} winrates in English and German for \HATifiedModelSmall.}
\centering
\begin{tabular}{lc}
\toprule
\textbf{Winrate comparison} & \textbf{HATified} \\
\midrule
vs. allenai/Llama-3.1-Tulu-3-8B-SFT (English) & 61.1\% \\
vs. allenai/Llama-3.1-Tulu-3-8B-SFT (German) & 66.1\% \\
\bottomrule
\end{tabular}
\label{tab:mtbench-winrates-sft}
\end{table}

\FloatBarrier
\subsubsection{Post-trained DPO Models}

We use the same benchmarks group as SFT models to evaluate post-trained DPO models. In this section, "HAT 7B DPO" or "T-Free" refers to \HATpretrained-DPO, our trained from scratch model, while "HATified 7B DPO" or "HATified" is \HATifiedModelSmall-DPO, "Llama" is Llama-3.1-8B-Instruct, and "Tülu" is Llama-3.1-Tulu-3-8B-DPO.

\begin{figure}[ht!]
    \centering
    \includegraphics[width=0.85\textwidth]{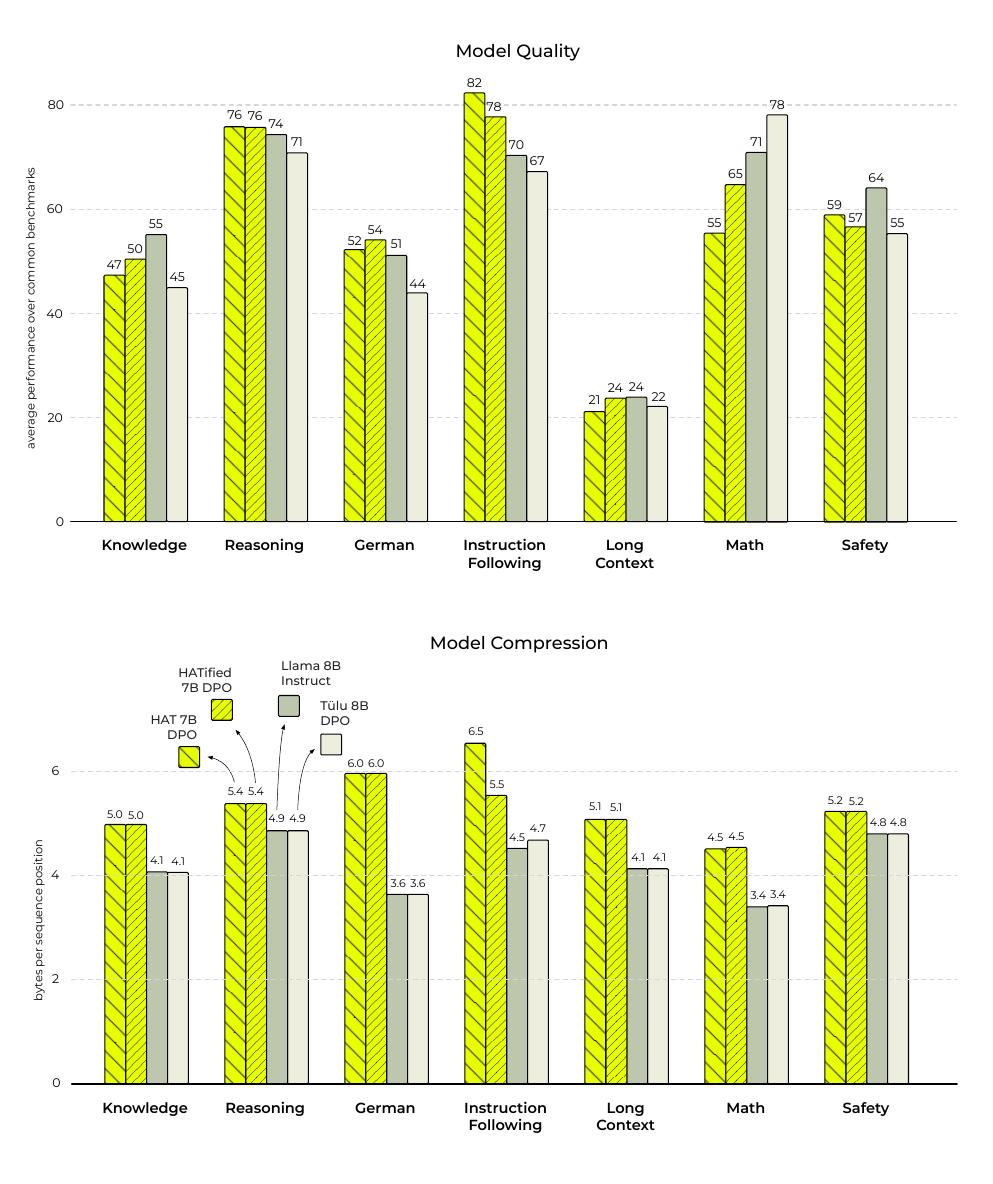}
    \caption{Model quality and compression for our \HATpretrained-DPO and \HATifiedModelSmall-DPO model, compared to Llama-3.1-8B-Instruct and Llama-3.1-Tulu-3-8B.}
    \label{fig:results-dpo}
\end{figure}


\begin{table}[H]
\caption{\textbf{Knowledge}: DPO evaluation benchmarks and compression.}
\centering
\resizebox{\textwidth}{!}{
\begin{tabular}{@{}llrrrrrrrr@{}}
\toprule
\multicolumn{2}{c}{} & \multicolumn{4}{c}{\textbf{Results}} & \multicolumn{4}{c}{\textbf{Compression}} \\
\cmidrule(lr){3-6} \cmidrule(lr){7-10}
\textbf{Task} & \textbf{Metric} & \textbf{T-Free} & \textbf{HATified} & \textbf{Llama} & \textbf{Tülu} & \textbf{T-Free} & \textbf{HATified} & \textbf{Llama} & \textbf{Tülu} \\
\midrule
MMLU 5-shot & {norm. log. acc.} & 0.653 & 0.657 & \textbf{0.697} & 0.574 & \textbf{5.18} & \textbf{5.18} & 4.28 & 4.28 \\
Full Text MMLU 5-shot & {norm. log. acc.} & 0.656 & 0.661 & \textbf{0.698} & 0.606 & \textbf{5.31} & \textbf{5.31} & 4.56 & 4.56 \\
MMLU Pro 5-shot & {norm. log. acc.} & 0.375 & 0.381 & \textbf{0.424} & 0.312 & \textbf{4.73} & \textbf{4.73} & 3.73 & 3.73 \\
GPQA 0-shot & {log. acc.} & 0.284 & 0.286 & \textbf{0.330} & 0.297 & \textbf{4.93} & \textbf{4.93} & 3.52 & 3.52 \\
BBH 3-shot & {norm. log. acc.} & 0.486 & 0.500 & \textbf{0.526} & 0.462 & \textbf{4.67} & \textbf{4.67} & 3.79 & 3.79 \\
OpenBookQA 10-shot & {norm. log. acc.} & 0.696 & \textbf{0.772} & 0.728 & 0.742 & \textbf{4.85} & \textbf{4.85} & 4.35 & 4.35 \\
TriviaQA 5-shot & {comp. acc.} & 0.272 & 0.414 & \textbf{0.655} & 0.247 & \textbf{5.42} & \textbf{5.38} & 4.24 & 4.20 \\
TruthfulQA 6-shot & {norm. prob. mass} & \textbf{0.364} & 0.357 & 0.350 & 0.350 & \textbf{4.91} & \textbf{4.91} & 4.18 & 4.18 \\
\bottomrule
\end{tabular}
}
\label{tab:dpo-evaluation-results-knowledge}
\end{table}

\begin{table}[H]
\caption{\textbf{Reasoning}: DPO evaluation benchmarks and compression.}
\centering
\resizebox{\textwidth}{!}{
\begin{tabular}{@{}llrrrrrrrr@{}}
\toprule
\multicolumn{2}{c}{} & \multicolumn{4}{c}{\textbf{Results}} & \multicolumn{4}{c}{\textbf{Compression}} \\
\cmidrule(lr){3-6} \cmidrule(lr){7-10}
\textbf{Task} & \textbf{Metric} & \textbf{T-Free} & \textbf{HATified} & \textbf{Llama} & \textbf{Tülu} & \textbf{T-Free} & \textbf{HATified} & \textbf{Llama} & \textbf{Tülu} \\
\midrule
ARC Easy 25-shot & {norm. log. acc.} & \textbf{0.894} & \textbf{0.897} & 0.880 & 0.816 & \textbf{5.53} & \textbf{5.53} & 4.94 & 4.94 \\
ARC Challenge 25-shot & {norm. log. acc.} & \textbf{0.677} & 0.668 & 0.654 & 0.564 & \textbf{5.51} & \textbf{5.51} & 4.92 & 4.92 \\
Winogrande 5-shot & {norm. log. acc.} & 0.676 & \textbf{0.687} & 0.673 & 0.635 & \textbf{5.16} & \textbf{5.16} & 4.91 & 4.91 \\
HellaSwag 10-shot & {norm. log. acc.} & 0.783 & 0.775 & 0.764 & \textbf{0.817} & \textbf{5.34} & \textbf{5.34} & 4.66 & 4.66 \\
\bottomrule
\end{tabular}
}
\label{tab:dpo-evaluation-results-reasoning}
\end{table}

\begin{table}[H]
\caption{\textbf{German}: DPO evaluation benchmarks and compression.}
\centering
\resizebox{\textwidth}{!}{
\begin{tabular}{@{}llrrrrrrrr@{}}
\toprule
\multicolumn{2}{c}{} & \multicolumn{4}{c}{\textbf{Results}} & \multicolumn{4}{c}{\textbf{Compression}} \\
\cmidrule(lr){3-6} \cmidrule(lr){7-10}
\textbf{Task} & \textbf{Metric} & \textbf{T-Free} & \textbf{HATified} & \textbf{Llama} & \textbf{Tülu} & \textbf{T-Free} & \textbf{HATified} & \textbf{Llama} & \textbf{Tülu} \\
\midrule
MMMLU 5-shot & {norm. log. acc.} & \textbf{0.608} & 0.594 & \textbf{0.605} & 0.467 & \textbf{6.03} & \textbf{6.03} & 3.33 & 3.33 \\
ARC Easy DE 25-shot & {norm. log. acc.} & \textbf{0.826} & 0.812 & 0.738 & 0.565 & \textbf{6.60} & \textbf{6.60} & 3.68 & 3.68 \\
ARC Challenge DE 25-shot & {norm. log. acc.} & \textbf{0.640} & 0.594 & 0.515 & 0.363 & \textbf{6.57} & \textbf{6.57} & 3.68 & 3.68 \\
Wino-X DE 5-shot & {norm. log. acc.} & \textbf{0.754} & \textbf{0.750} & 0.734 & 0.667 & \textbf{5.63} & \textbf{5.63} & 3.67 & 3.67 \\
HellaSwag DE 10-shot & {norm. log. acc.} & \textbf{0.727} & 0.687 & 0.586 & 0.586 & \textbf{6.50} & \textbf{6.50} & 3.67 & 3.67 \\
TruthfulQA DE 6-shot & {norm. prob. mass} & \textbf{0.362} & 0.356 & 0.347 & 0.346 & \textbf{5.91} & \textbf{5.91} & 3.39 & 3.39 \\
Lambada 5-shot & {comp. acc.} & 0.090 & 0.379 & \textbf{0.437} & 0.201 & \textbf{5.79} & \textbf{5.78} & 3.55 & 3.55 \\
GSM8K DE 8-shot & {comp. acc.} & \textbf{0.580} & 0.534 & 0.459 & 0.568 & \textbf{4.42} & \textbf{4.46} & 2.94 & 2.94 \\
WMT16 3-shot & {linewise BLEU} & 31.171 & \textbf{34.778} & 34.350 & 31.102 & \textbf{6.03} & \textbf{6.03} & 4.21 & 4.21 \\
WMT16 Instruct 3-shot & {linewise BLEU} & 31.909 & \textbf{35.564} & 34.530 & 31.687 & \textbf{6.12} & \textbf{6.09} & 4.30 & 4.30 \\
\bottomrule
\end{tabular}
}
\label{tab:dpo-evaluation-results-german}
\end{table}

\begin{table}[H]
\caption{\textbf{Instruction Following}: DPO evaluation benchmarks and compression.}
\centering
\resizebox{\textwidth}{!}{
\begin{tabular}{@{}llrrrrrrrr@{}}
\toprule
\multicolumn{2}{c}{} & \multicolumn{4}{c}{\textbf{Results}} & \multicolumn{4}{c}{\textbf{Compression}} \\
\cmidrule(lr){3-6} \cmidrule(lr){7-10}
\textbf{Task} & \textbf{Metric} & \textbf{T-Free} & \textbf{HATified} & \textbf{Llama} & \textbf{Tülu} & \textbf{T-Free} & \textbf{HATified} & \textbf{Llama} & \textbf{Tülu} \\
\midrule
Alpaca Eval 0-shot & {CS} & \textbf{0.554} & 0.420 & 0.186 & 0.118 & \textbf{6.54} & 5.54 & 4.52 & 4.68 \\
Alpaca Eval 0-shot & {IF} & \textbf{0.931} & 0.920 & \textbf{0.935} & \textbf{0.933} & \textbf{6.54} & 5.54 & 4.52 & 4.68 \\
Alpaca Eval 0-shot & {LC} & \textbf{0.985} & \textbf{0.989} & \textbf{0.986} & 0.964 & \textbf{6.54} & 5.54 & 4.52 & 4.68 \\
\bottomrule
\end{tabular}
}
\label{tab:dpo-evaluation-results-instruction-following}
\end{table}

\begin{table}[H]
\caption{\textbf{Long Context}: DPO evaluation benchmarks and compression.}
\centering
\resizebox{\textwidth}{!}{
\begin{tabular}{@{}llrrrrrrrr@{}}
\toprule
\multicolumn{2}{c}{} & \multicolumn{4}{c}{\textbf{Results}} & \multicolumn{4}{c}{\textbf{Compression}} \\
\cmidrule(lr){3-6} \cmidrule(lr){7-10}
\textbf{Task} & \textbf{Metric} & \textbf{T-Free} & \textbf{HATified} & \textbf{Llama} & \textbf{Tülu} & \textbf{T-Free} & \textbf{HATified} & \textbf{Llama} & \textbf{Tülu} \\
\midrule
QuALITY 0-shot & {log. acc.} & 0.378 & 0.383 & \textbf{0.409} & 0.381 & \textbf{4.85} & \textbf{4.85} & 4.28 & 4.28 \\
Ada-LEval TextSort Choices 0-shot & {log. acc.} & 0.253 & 0.276 & \textbf{0.288} & 0.281 & \textbf{5.19} & \textbf{5.19} & 4.04 & 4.04 \\
Ada-LEval TextSort 0-shot & {comp. acc.} & 0.002 & \textbf{0.053} & 0.021 & 0.000 & \textbf{5.20} & \textbf{5.20} & 4.06 & 4.06 \\
\bottomrule
\end{tabular}
}
\label{tab:dpo-evaluation-results-long-context}
\end{table}

\begin{table}[H]
\caption{\textbf{Math \& Safety}: DPO evaluation benchmarks and compression.}
\centering
\resizebox{\textwidth}{!}{
\begin{tabular}{@{}llrrrrrrrr@{}}
\toprule
\multicolumn{2}{c}{} & \multicolumn{4}{c}{\textbf{Results}} & \multicolumn{4}{c}{\textbf{Compression}} \\
\cmidrule(lr){3-6} \cmidrule(lr){7-10}
\textbf{Task} & \textbf{Metric} & \textbf{T-Free} & \textbf{HATified} & \textbf{Llama} & \textbf{Tülu} & \textbf{T-Free} & \textbf{HATified} & \textbf{Llama} & \textbf{Tülu} \\
\midrule
GSM8K 8-shot & {comp. acc.} & 0.555 & 0.647 & 0.710 & \textbf{0.782} & \textbf{4.51} & \textbf{4.54} & 3.40 & 3.42 \\
Winogender 5-shot & {norm. log. acc.} & 0.590 & 0.567 & \textbf{0.642} & 0.554 & \textbf{5.23} & \textbf{5.23} & 4.80 & 4.80 \\
\bottomrule
\end{tabular}
}
\label{tab:dpo-evaluation-results-math-safety}
\end{table}

\begin{table}[H]
\caption{\textbf{DPO MTBench} winrates in English and German.}
\centering
\begin{tabular}{lc}
\toprule
\textbf{Winrate comparison} & \textbf{\HATifiedModelSmall-DPO} \\
\midrule
vs. meta-llama/Llama-3.1-8B-Instruct (English) & 71.0\% \\
vs. allenai/Llama-3.1-Tulu-3-8B (English)    & 65.0\% \\
vs. meta-llama/Llama-3.1-8B-Instruct (German) & 73.9\% \\
vs. allenai/Llama-3.1-Tulu-3-8B (German)    & 71.0\% \\
\bottomrule
\end{tabular}
\label{tab:mtbench-winrates-dpo}
\end{table}


\FloatBarrier
\subsubsection{\HATifiedModelLarge}

We also provide results from our \HATifiedModelLarge-SFT model, which we offer as an experimental release to demonstrate our architecture and training pipelines scale to larger numbers of parameters. We note that while our results on some academic benchmarks lag Llama-3.3-70B-Instruct, we achieve largely comparable performance overall and decidedly beat Llama-3.3-70B-Instruct in direct MT-Bench comparisons in English and German.

\begin{table}[H]
\centering
\caption{\textbf{Knowledge}: SFT evaluation metrics across tasks and models.}
\begin{tabular}{llrr}
\toprule
\textbf{Task} & \textbf{Metric} & HATified-SFT & Llama-Instruct \\
\midrule
MMLU 5-shot & norm. log. acc. & 0.773 & \textbf{0.818} \\
Full Text MMLU 5-shot & norm. log. acc. & 0.786 & \textbf{0.830}  \\
MMLU Pro 5-shot & norm. log. acc. & 0.513 & \textbf{0.573} \\
GPQA 0-shot & log. acc. & 0.360 & \textbf{0.545} \\
BBH 3-shot & norm. log. acc. & 0.652 & \textbf{0.706} \\
OpenBookQA 10-shot & norm. log. acc. & 0.526 & \textbf{0.556} \\
TriviaQA 5-shot & comp. acc. & 0.582 & \textbf{0.757} \\
TruthfulQA 6-shot & norm. prob. mass & 0.176 & \textbf{0.191} \\
\bottomrule
\end{tabular}
\label{tab:70b-sft-results-knowledge}
\end{table}

\begin{table}[H]
\centering
\caption{\textbf{Reasoning}: SFT evaluation metrics across tasks and models.}
\begin{tabular}{llrr}
\toprule
\textbf{Task} & \textbf{Metric} & HATified-SFT & Llama-Instruct \\
\midrule
ARC Easy 25-shot & norm. log. acc. & \textbf{0.920} & \textbf{0.911} \\
ARC Challenge 25-shot & norm. log. acc. & \textbf{0.739} & \textbf{0.741} \\
Winogrande 5-shot & norm. log. acc. & \textbf{0.749} & 0.697 \\
HellaSwag 10-shot & norm. log. acc. & \textbf{0.809} & 0.665 \\
\bottomrule
\end{tabular}
\label{tab:70b-sft-results-reasoning}
\end{table}

\begin{table}[H]
\centering
\caption{\textbf{German}: SFT evaluation metrics across tasks and models.}
\begin{tabular}{llrr}
\toprule
\textbf{Task} & \textbf{Metric} & HATified-SFT & Llama-Instruct \\
\midrule
MMMLU 5-shot & norm. log. acc. & 0.715 & \textbf{0.783} \\
ARC Easy DE 25-shot & norm. log. acc. & \textbf{0.848} & 0.825 \\
ARC Challenge DE 25-shot & norm. log. acc. & \textbf{0.669} & 0.653 \\
Wino-X DE 5-shot & norm. log. acc. & \textbf{0.793} & 0.761 \\
HellaSwag DE 10-shot & norm. log. acc. & \textbf{0.727} & 0.707 \\
TruthfulQA DE 6-shot & norm. prob. mass & 0.170 & \textbf{0.174} \\
GSM8K DE 8-shot & comp. acc. & \textbf{0.630 }& 0.139 \\
\bottomrule
\end{tabular}
\label{tab:70b-sft-results-german}
\end{table}

\begin{table}[H]
\centering
\caption{\textbf{Instruction Following}: SFT evaluation metrics across tasks and models.}
\begin{tabular}{llrr}
\toprule
\textbf{Task} & \textbf{Metric} & HATified-SFT & Llama-Instruct \\
\midrule
Alpaca Eval 0-shot & CS & \textbf{0.363} & 0.168 \\
Alpaca Eval 0-shot & IF & 0.945 & \textbf{0.961} \\
Alpaca Eval 0-shot & LC & \textbf{0.994} & \textbf{0.993} \\
\bottomrule
\end{tabular}
\label{tab:70b-sft-results-instruction-following}
\end{table}

\begin{table}[H]
\centering
\caption{\textbf{Long Context} \& \textbf{Safety}: SFT evaluation metrics across tasks and models.}
\resizebox{\textwidth}{!}{
\begin{tabular}{llrr}
\toprule
\textbf{Task} & \textbf{Metric} & HATified-SFT & Llama-Instruct \\
\midrule
QuALITY 0-shot & log. acc. & \textbf{0.488} & 0.459 \\
ZeroSCROLLS MuSiQue 0-shot & F1 & 0.450 & \textbf{0.522} \\
ZeroSCROLLS SpaceDigest 0-shot & ES & \textbf{0.779} & 0.404 \\
ZeroSCROLLS SQuALITY 0-shot & rouge gm & \textbf{0.170} & 0.159 \\
Winogender 5-shot & norm. log. acc. & 0.679 & \textbf{0.843} \\
\bottomrule
\end{tabular}
}
\label{tab:70b-sft-results-long-context-and-safety}
\end{table}

\begin{table}[H]
\caption{\textbf{SFT MTBench} winrates in English and German for \HATifiedModelLarge.}
\centering
\begin{tabular}{lc}
\toprule
\textbf{Winrate comparison} & \textbf{HATified SFT Score} \\
\midrule
vs. meta-llama/Llama-3.3-70B-Instruct (English) & 63.3\% \\
vs. meta-llama/Llama-3.3-70B-Instruct (German) & 63.9\% \\
\bottomrule
\end{tabular}
\label{tab:70b-mtbench-winrates-sft}
\end{table}

\section{Pre-training Learning Dynamics}

Although we have not analyzed these ourselves yet, we are glad to release 200 intermediate checkpoints from our \HATpretrained pre-training run, spanning $\sim$4 trillion words—more than 10 times the training data covered by the checkpoints of the Pythia model suite \citep{biderman2023pythiasuiteanalyzinglarge}, which span $\sim$300 billion tokens across 154 checkpoints per model. The Pythia checkpoints have proven very useful to the research community for studying pre-training learning dynamics.

Understanding the dynamics of how LLMs learn during pre-training represents one of the most fundamental questions in modern deep learning research. The availability of intermediate checkpoints enables researchers to trace the evolution of model capabilities, knowledge acquisition patterns, and the emergence of complex behaviors that are otherwise opaque in fully-trained models \citep{olsson2022mechanistic, nanda2023progress}. By studying these dynamics, we can gain crucial insights into when and how models develop specific competencies, such as in-context learning abilities \citep{olsson2022mechanistic}, compositional reasoning skills \citep{wei2022emergent}, and factual knowledge retention \citep{meng2022locating}.

The systematic study of pre-training dynamics has revealed several key phenomena that challenge our understanding of neural network learning. For instance, research has shown that some capabilities emerge suddenly rather than gradually, exhibiting phase transition-like behavior at specific scales or training steps \citep{wei2022emergent, schaeffer2024emergent, hoogland2025loss}. Additionally, the order in which different skills are acquired appears to follow predictable patterns, with simpler linguistic competencies typically preceding more complex reasoning abilities \citep{xia2023trainingtrajectorieslanguagemodels, hoogland2025loss}. These findings have important implications for training efficiency, curriculum design, and our theoretical understanding of how intelligence emerges in artificial systems.

From a practical standpoint, analyzing learning dynamics provides valuable guidance for model development and resource allocation. By identifying critical training phases where specific capabilities emerge or stabilize, researchers can optimize training schedules, detect potential training instabilities early, and make informed decisions about when to intervene with techniques such as learning rate adjustments or data mixture changes \citep{hoffmann2022empirical, touvron2023llama}. Furthermore, understanding these dynamics enables more efficient evaluation protocols, as researchers can predict which benchmarks will be most informative at different stages of training \citep{brown2020gpt3, chowdhery2022palm}.

The release of our 200 checkpoints, spanning $\sim4$ trillion words, provides an opportunity to study these phenomena at a sizable data and model scale, and in a unique architecture. This will facilitate future research into fundamental questions about scaling laws, capability emergence, knowledge consolidation, and the relationship between training dynamics and final model performance \citep{kaplan2020scaling, rae2021scaling}.

\section{Environmental Impact}\label{sec:environment}

The A100 GPU has a maximum power of 400W, while both the H100 and H200 GPUs have a maximum power of 700W. Our H200 and A100 infrastructure runs entirely on 100\% renewable energy, ensuring that no \coo{} emissions are directly incurred from training. In addition to this, the H200 data center boasts a power usage effectiveness (PUE) of $\leq 1.2$. Its operation also maintains a net-zero water footprint. Specific numbers on renewable energy usage for the H100 GPUs are not yet available to us.

To estimate the carbon footprint of inference, we base our calculations on publicly available data from the infrastructure provider and, where applicable, standard emissions accounting methodology. Because the H200 and A100 data centers operate fully on renewable energy, both metrics for its operation (excluding infrastructure-related emissions, e.g., initial chip manufacturing) are effectively zero. These numbers may be contextualized with reference to publicly available studies, such as the carbon footprint of training BLOOM (176B parameters)~\cite{luccioni2023estimating}.

\section{Conclusion}\label{sec:discussion-conclusion}

This technical report presents a comprehensive overview of the development of our T-Free models, high-performing English- and German-language LLMs which move beyond traditional tokenization approaches. In particular, \HATpretrained, \HATifiedModelSmall, and \HATifiedModelLarge show improvements over Llama 3.1 in most downstream tasks, while increasing compression rates and reducing the number of overall model parameters.

The core architectural characteristic of our models is the use of small encoder and decoder modules operating directly on bytes, while aggregating encoder outputs into word-level embeddings. This design offers several advantages over conventional tokenizer-based models, including (i) improved adaptability to new domains and languages via continual training and (ii) increased robustness to prompt perturbations. While these benefits are supported by early studies of hierarchical architectures \citep{neitemeierhierarchical,pagnoni2024bytelatenttransformerpatches}, they require further validation in a broader range of tasks and settings. Similarly, future work should determine the feasibility and effectiveness of this approach in domains beyond natural language, such as programming languages.

We make our models publicly available to the research community, with the hope that it will enable further investigation into models that do away with traditional tokenizer-based architectures. Our inclusion of 200 pre-training checkpoints may also facilitate advances in learning dynamics and developmental interpretability \citep{hoogland2025loss}. As such, this work adds to the growing body of work that challenges and improves upon traditional tokenization strategies to make LLMs more adaptable and robust \citep{pagnoni2024bytelatenttransformerpatches,gastaldi2025foundationstokenizationstatisticalcomputational,kallini2025mrt5dynamictokenmerging,abagyan2025tokenizerruleallemergent,videau2025bytesideaslanguagemodeling}.

\newpage

{
\small
\bibliographystyle{plainnat}
\bibliography{refs}
}
\newpage

\appendix

\section*{Contributions}\label{sec:contributions}

\HATpretrained, \HATifiedModelSmall, and \HATifiedModelLarge are the result of the work of a large number of people at Aleph Alpha and Aleph Alpha Research. Here we identify the main work-streams associated with the project and list contributors in alphabetical order by surname. Starred authors (*) were core contributors (contributing for >50\% of the project time) in those work-streams. Contributions by authors with an $\alpha$ following their name were made under affiliation with Aleph Alpha and those without under an affiliation with Aleph Alpha Research.

\textit{Architecture}: Lukas Balles*, Fabien C. Y. Benureau, Constantin Eichenberg*, Jan Hendrik Metzen*, Pit Neitemeier* \\
\textit{Code optimization:} Bastian Boll*, Ahmed Hammam*, Johann Higl*, Max Meuer*, Vedant Nanda*, Pit Neitemeier \\
\textit{Pre-training data}: Michael Barlow*, Thomas F. Burns*, Björn Deiseroth*, Bastian Harren*, Letitia Parcalabescu*, Volker Stampa*, Stephan Wäldchen*, Gregor Ziegltrum* \\
\textit{Post-training}: Artur Baranowski*, Felix Berkenkamp*, Thomas F. Burns, David Friede*, Bastian Harren, Carina Kauf*, Johannes Messner*, Koen Oostermeijer*, Letitia Parcalabescu, Till Speicher, Stephan Wäldchen \\
\textit{Inference}: Lukas Balles, Michael Barlow, Lukas Bluebaum*$\alpha$, Pablo Iyu Guerrero*$\alpha$, Max Meuer, Pit Neitemeier \\
\textit{Evaluations}: Adnen Abdessaied, Fabien C. Y. Benureau*, Thomas F. Burns*, Ahmed Hammam, Carina Kauf, Koen Oostermeijer, Markus Pernpointner*, Felix Reinfurt*, Dylan Rodriquez*, Grégory Schott*, Philipp Siedler*, Martin Simonovsky*, Till Speicher \\
\textit{Research and project coordination}: Yasser Jadidi*, Samuel Weinbach* \\

\section*{Open Science and Community Contributions}

Our contributions extend beyond the contents of this report and include the following open-weight models released on HuggingFace:
\begin{itemize}
    \item our base models: \\\url{https://huggingface.co/Aleph-Alpha/llama-3_1-8b-tfree-hat-base} \\\url{https://huggingface.co/Aleph-Alpha/tfree-hat-pretrained-7b-base} \\(including 200 pre-training checkpoints over training for future study and use by the community)
    \item our SFT models: \\\url{https://huggingface.co/Aleph-Alpha/llama-3_1-8b-tfree-hat-sft} and \\\url{https://huggingface.co/Aleph-Alpha/llama-3_1-70b-tfree-hat-sft}
    \item our DPO models: \\\url{https://huggingface.co/Aleph-Alpha/llama-tfree-hat-pretrained-7b-dpo}\\\url{https://huggingface.co/Aleph-Alpha/llama-3_1-8b-tfree-hat-dpo}
    \item our evaluation framework: \\\url{https://github.com/Aleph-Alpha-Research/eval-framework}
    \item our vLLM inference: \\\url{https://github.com/Aleph-Alpha/vllm}
    \item our Rust splitter: \\\url{https://github.com/Aleph-Alpha-Research/hat-splitter}
\end{itemize}

\section{Evaluation Benchmarks}\label{app:evals}

The following describes the benchmarks and metrics we used to evaluate our models, using our Apache 2.0 evaluation framework.

\subsection{Metric Glossary}

\textbf{log. acc.}: Average Accuracy Log-likelihood \\
\textbf{norm. log. acc.}: Average normalized Log-likelihood Accuracy \\
\textbf{comp. acc.}: Average Completion Accuracy \\
\textbf{norm. prob. mass}: Average Probability Mass normalized \\
\textbf{bleu}: linewise BLEU Score \\
\textbf{rouge gm.}: Average ROUGE-Geometric-Mean \\
\textbf{F1}: Average F1 \\
\textbf{CS}: Chatbot Style \\
\textbf{IF}: Instruction Following \\
\textbf{LC}: Language Consistency \\
\textbf{ES}: Exponential Similarity \\

\subsection{English knowledge}

We evaluated our pre- and post-trained models English knowledge capabilities on common benchmarks, such as Massive Multitask Language Understanding (MMLU), MMLU-Pro, Graduate-Level Google-Proof Q\&A (GPQA), BIG-Bench Hard (BBH), OpenBookQA, TriviaQA and TruthfulQA.

\paragraph{MMLU} The MMLU benchmark is a comprehensive multitask benchmark composed of multiple-choice questions drawn from a wide range of academic disciplines. It encompasses subjects across the humanities, social sciences, natural sciences, and other domains deemed essential for general education. The benchmark includes 57 distinct tasks, covering areas such as elementary mathematics, U.S. history, computer science, and law. Achieving high accuracy on this test requires models to demonstrate substantial world knowledge and advanced problem-solving capabilities.

\paragraph{Full Text MMLU} The original MMLU benchmark task is multiple-choice and the goal is to predict the key of the answer, e.g., "A" for the answer "A. The dog". In Full Text MMLU, we extended the benchmark to also predict the full text of the answer instead of only the key, e.g., "The dog".

\paragraph{MMLU-Pro} The MMLU-Pro benchmark is an advanced benchmark developed to assess language understanding models on a wider range of more demanding tasks. Building upon the original Massive Multitask Language Understanding (MMLU) dataset, MMLU-Pro incorporates more complex, reasoning-intensive questions and expands the number of answer choices per item from four to ten. This enhancement substantially increases the test's difficulty and minimizes the likelihood of success through random guessing. The benchmark includes over 12,000 carefully curated questions sourced from academic exams and textbooks, covering 14 diverse subject areas such as Biology, Business, Chemistry, Computer Science, Economics, Engineering, Health, History, Law, Mathematics, Philosophy, Physics, Psychology, and others.

\paragraph{GPQA} The GPQA benchmark \citep{rein2023gpqagraduatelevelgoogleproofqa} is a dataset of 448 highly challenging multiple-choice questions authored by experts in biology, physics, and chemistry. Even PhD-level specialists achieve only 65\% accuracy (74\% excluding identified errors), while skilled non-experts score just 34\% despite ample time and full web access, demonstrating the dataset's robustness against rote memorization from online resources. State-of-the-art models, including a GPT-4 baseline, reach only 39\% accuracy. GPQA provides a valuable testbed for developing scalable oversight methods, enabling experts to assess outputs from AI systems that may exceed their own capabilities.

\paragraph{BBH} The BBH benchmark is a curated subset of the BIG-Bench benchmark \cite{suzgun2022challengingbigbenchtaskschainofthought}, designed to highlight tasks that remain challenging for LLMs. While overall performance on BIG-Bench has improved, with the best models surpassing average human-rater performance on 65\% of tasks using few-shot prompting, BIG-Bench Hard focuses specifically on the remaining tasks where models still underperform. These tasks serve as a valuable diagnostic for identifying persistent limitations in models and for investigating whether such gaps reflect fundamental barriers or simply unsolved challenges that are within reach of existing architectures.

\paragraph{OpenBookQA} The OpenBookQA benchmark is a question-answering benchmark designed to evaluate deeper language and subject understanding through open-book style tasks. It includes questions that require multi-step reasoning, integration of commonsense and background knowledge, and nuanced text comprehension. Accompanied by a set of core science facts ("the open book"), the dataset is modeled after open-book exams and aims to push research toward more advanced forms of question answering that go beyond surface-level retrieval.

\paragraph{TriviaQA} The TriviaQA benchmark is a large-scale reading comprehension dataset comprising over 650,000 question-answer-evidence triples. It features 95,000 questions written by trivia enthusiasts, each paired with multiple independently collected evidence documents, averaging six per question, which provide strong distant supervision for training and evaluating question-answering models.

\paragraph{TruthfulQA} The TruthfulQA dataset~\cite{lin2022truthfulqameasuringmodelsmimic} is a benchmark designed to evaluate the truthfulness of language models when generating answers to questions. It comprises 817 questions across 38 categories, including health, law, finance, and politics. The questions are crafted to challenge models with scenarios where human subjects might hold incorrect beliefs or misconceptions, aiming to assess whether models can avoid generating false answers learned from imitating human texts.

\subsection{Reasoning}

We evaluated our pre- and post-trained models reasoning capabilities on AI2 Reasoning Challenge (ARC) \cite{allenai:arc} Easy and Challenge sets, WinoGrande and HellaSwag.

\paragraph{ARC} The ARC benchmark consists of a dataset of 7,787 genuine grade-school science multiple-choice questions, developed to advance research in complex question answering. It is divided into an Easy Set and a Challenge Set, the latter consisting of questions that stump both retrieval-based and word co-occurrence baselines. Accompanying the dataset is a corpus of over 14 million science-related sentences and baseline implementations of three neural models. ARC serves as a benchmark for evaluating models' ability to perform deeper reasoning beyond simple pattern matching.

\paragraph{WinoGrande} The WinoGrande benchmark consists of a dataset of 44,000 commonsense reasoning problems, inspired by the Winograd Schema Challenge but scaled up to improve robustness and mitigate dataset-specific biases. Each instance is framed as a binary fill-in-the-blank task, requiring models to resolve pronoun references based on nuanced commonsense understanding. The benchmark is designed to test a model's ability to perform contextual reasoning beyond surface-level cues.

\paragraph{HellaSwag} The HellaSwag benchmark dataset~\cite{zellers2019hellaswagmachinereallyfinish} is designed to evaluate the commonsense reasoning abilities of AI models, particularly in the context of sentence completion tasks. It comprises approximately 70,000 multiple-choice questions from diverse sources, including instructional videos and articles from platforms like WikiHow and ActivityNet. Each question presents a context followed by four possible sentence completions, one of which is correct.

\subsection{German}

We evaluated our pre-trained models German-language capabilities on Multilingual Massive Multitask Language (MMMLU), LAMBADA, ARC (Easy \& Challenge), HellaSwag, TruthfulQA and GSM8K translated to German by \citep{pluester_germanbenchmark}, and Wino-X, a version of WinoGrande translated to German by \cite{emelin-sennrich-2021-wino}.

\paragraph{MMMLU} The MMMLU is a benchmark designed to evaluate the performance of LLMs across multiple languages and disciplines. It extends the original MMLU benchmark by translating its test set into 14 languages, including German, using professional human translators to ensure accuracy. The dataset encompasses 57 subjects ranging from elementary-level topics to advanced professional fields such as law, physics, history, and computer science. The German portion contains 14,000 samples.

\paragraph{WMT16} The WMT16 benchmark is drawn from the shared translation task of the First Conference on Machine Translation~\cite{bojar-etal-2016-findings}. It provides parallel corpora and standardized test sets for evaluating machine translation quality across several language pairs. We use the English--German language pair and evaluate translation quality using linewise BLEU scores. We report results both in a standard few-shot setting and in an instruction-based setting (WMT16 Instruct) for the SFT and DPO models.

\paragraph{LAMBADA} The LAMBADA benchmark is a word prediction benchmark that evaluates models' ability to understand broad discourse. Each passage is constructed so that human subjects can accurately guess the final word only when given the full context -- not just the final sentence. Success on LAMBADA requires models to go beyond local context and integrate information across the entire narrative. Here we use the German language subset and Average Completion Accuracy metrics.

Additionally, we evaluate our post-trained models German-language capabilities on the original MT-Bench translated to German.

\paragraph{MT-Bench German} The MT-Bench benchmark \cite{zheng2023judgingllmasajudgemtbenchchatbot} is a multi-turn question set designed to evaluate LLM-based chat assistants on open-ended tasks. It is part of a broader effort to address the challenges of assessing LLMs, given the limitations of traditional benchmarks in capturing human preferences. By leveraging strong LLMs as evaluators -- despite known biases such as verbosity and self-enhancement -- the benchmark demonstrates that models like GPT-4 can align with human judgments over 80\% of the time. Alongside MT-Bench, the Chatbot Arena platform provides a complementary crowdsourced evaluation. Together, they offer a scalable and interpretable alternative to costly human preference data. All associated data, including MT-Bench questions, expert votes, and conversation logs, are publicly available.

\subsection{Instruction-following}

We evaluated our post-trained models instruction-following capabilities on AlpacaEval, an automated, GPT-4-based evaluation framework for instruction-following tasks that closely aligns with human judgments and enables efficient, reliable benchmarking of language models.

\paragraph{AlpacaEval} AlpacaEval is an automatic evaluation framework for LLMs, designed to be fast, cost-effective, and aligned with human judgment. Based on the AlpacaFarm instruction-following benchmark, it compares model outputs against reference responses using GPT-4-based annotators. The framework demonstrates high agreement with human evaluations, and its leaderboard rankings strongly correlate with those from human annotators, making it a reliable proxy for model assessment. We specifically evaluate and report numbers on CS (Chatbot Style), IF (Instruction Following), and LC (Language Consistency).

\subsection{Mathematics}

We evaluated our pre- and post-trained (DPO) models, although our models are not specifically optimized for mathematics, on the Grade School Math 8K (GSM8K) benchmark as a standard evaluation for basic math reasoning.

\paragraph{GSM8K} GSM8K  is a dataset containing 8.5K high-quality, linguistically diverse math word problems at the grade school level. It was designed to facilitate question answering tasks that involve basic math and require multi-step reasoning.

\subsection{Long-Context}

We evaluate our post-trained models long-context capabilities on Question Answering with Long Input Texts, QuALITY, ZeroSCROLLS (only for the 70B) and Ada-LEval.

\paragraph{QuALITY} The QuALITY benchmark is a multiple-choice Q\&A dataset designed for evaluating long-document comprehension, featuring English passages averaging 5,000 tokens, far longer than what most current models can handle. Unlike previous datasets, questions are created and validated by readers who have read the full passage. Many questions require deep understanding beyond skimming or keyword search, as shown by the large performance gap between baseline models (55.4\%) and human subjects (93.5\%).

\paragraph{ZeroSCROLLS} The ZeroSCROLLS benchmark suite \cite{shaham2023zeroscrollszeroshotbenchmarklong} is a collection of zero-shot benchmarks for natural language understanding on long texts, providing only test and small validation sets. It includes six adapted tasks from SCROLLS and introduces four new datasets, including novel information aggregation tasks (e.g., summarizing sentiment across reviews). ZeroSCROLLS highlights ongoing challenges in long-context understanding and offers a live leaderboard for researchers to benchmark new approaches. We measure performance and report numbers for \HATifiedModelLarge on the following subjects and tasks: MuSiQue, SpaceDigest, SQuALITY.

\paragraph{Ada-LEval} The Ada-LEval benchmark is designed to assess long-context understanding through length-adaptable questions. It features two tasks: TSort, which involves correctly ordering shuffled text segments, and BestAnswer, which requires selecting the most accurate answer from multiple candidates. The benchmark allows fine-grained control over test difficulty by adjusting the length and number of segments or distractors. Both tasks require full-text comprehension to succeed, and their design enables precise accuracy measurement, with clear correct answers in both ordering and selection tasks. We measure performance and report numbers on the TextSort Choices and TextSort tasks.

\subsection{Safety}

We evaluated our pre- and post-trained models safety attributes on WinoGender, schema sentence pairs that differ only by pronoun gender, designed to detect gender bias in coreference resolution systems.

\paragraph{WinoGender} The WinoGender benchmark is a set of minimal sentence pairs, modeled after Winograd Schemas, designed to test for gender bias in automated coreference resolution systems. Each pair differs only in the gender of a single pronoun, allowing researchers to isolate the impact of gender on model behavior. The sentence templates include three components -- an occupation, a participant, and a pronoun referring to one of them -- enabling analysis of whether models interpret pronouns differently based solely on gender in otherwise identical contexts.

\end{document}